\ifdefined\XeTeXversion
\else
  \ifdefined\pdfoutput\pdfoutput=1\fi
\fi
\documentclass[11pt,a4paper]{article}

\usepackage[T1]{fontenc}
\usepackage[utf8]{inputenc}
\usepackage{lmodern}
\usepackage{microtype}
\usepackage[a4paper,left=31mm,right=25mm,top=24mm,bottom=25mm]{geometry}
\usepackage{graphicx}
\usepackage{booktabs}
\usepackage{multirow}
\usepackage{tabularx}
\usepackage{longtable}
\usepackage{amsmath,amssymb,amsfonts}
\usepackage{enumitem}
\usepackage[numbers,sort&compress]{natbib}
\usepackage{xcolor}
\usepackage{xspace}
\usepackage[section]{placeins}
\usepackage{caption}
\usepackage{url}
\usepackage{hyperref}

\hypersetup{
  hidelinks,
  pdftitle={XHotpotQA: A Benchmark for Cross-Lingual Knowledge Composition in Multi-Hop Question Answering},
  pdfauthor={Iman Barati; Arash Ghafouri; Behrouz Minaei-Bidgoli},
  pdfsubject={Cross-lingual multi-hop question answering benchmark and evaluation}
}

\newcolumntype{Y}{>{\raggedright\arraybackslash}X}
\newcommand{\dataset}{\mbox{XHotpotQA}\xspace}
\newcommand{\datasetplus}{\mbox{XHotpotQA+}\xspace}
\newcommand{\xreader}{\mbox{xLlama-Reader}\xspace}
\newcommand{\xselector}{\mbox{xLlama-Selector}\xspace}

\newcommand{\ind}{\operatorname{I}}
\definecolor{XHDeepBlue}{HTML}{164E63}
\definecolor{XHSky}{HTML}{38BDF8}
\definecolor{XHPale}{HTML}{F0F9FF}
\definecolor{XHGold}{HTML}{B45309}
\definecolor{XHGreen}{HTML}{047857}

\newenvironment{promptbox}[1]{%
  \par\medskip\noindent\begingroup
  \textcolor{XHDeepBlue}{\sffamily\bfseries #1}\par\smallskip
  \begin{quote}\small\ttfamily\raggedright
}{%
  \end{quote}\endgroup\par\medskip
}

\newenvironment{examplebox}[2]{%
  \par\smallskip\noindent\begingroup
  \textcolor{#2}{\sffamily\bfseries #1}\par\smallskip
  \begin{quote}\small\raggedright
}{%
  \end{quote}\endgroup\par\smallskip
}

\title{\textbf{XHotpotQA: A Benchmark for Cross-Lingual Knowledge Composition in Multi-Hop Question Answering}}
\author{Iman Barati \quad Arash Ghafouri \quad Behrouz Minaei-Bidgoli\\[0.65em]
\small School of Computer Engineering, Iran University of Science and Technology\\
\small Tehran 16846-13114, Iran\\
\small Corresponding author: \texttt{b\_minaei@iust.ac.ir}}
\date{}

\begin{document}
\maketitle

\begin{abstract}
Knowledge-intensive multi-hop question answering requires systems to select evidence and compose dependent facts, yet multilingual benchmarks usually translate an entire example into one language. This hides failures at language boundaries inside the reasoning chain. We introduce \dataset, a controlled benchmark for cross-lingual knowledge composition over mixed-language evidence. Each instance is modeled as an evidence-dependency graph whose question, bridge evidence, answer-bearing evidence, and distractors have explicit language assignments. The audited resource contains 15,661 training and 7,405 validation instances, with sentence-level support supervision and supplied distractors. In validation, 99.81\% of items cross the question--gold-evidence language interface and 95.60\% use gold paragraphs in different languages. Across three reader artifacts, full question--evidence mismatch is associated with 10.25--15.79 lower Unicode-aware answer F1 than partial alignment, and different-script evidence with deficits of 11.98--23.70 points; the corresponding adapted-selector contrasts are 1.71 and 1.78 points. Under this supplied-candidate design, the evaluated readers therefore show substantially larger condition-associated deficits than the selector. \dataset\ provides role-aware diagnostics, modular evaluation, and an audited test bed for knowledge-based systems that must integrate evidence across languages.
\end{abstract}

\noindent\textbf{Keywords:} cross-lingual question answering; multi-hop reasoning; knowledge composition; evidence graphs; multilingual benchmark; language-model evaluation
\medskip

\section{Introduction}\label{sec:introduction}

Knowledge-based question-answering systems must do more than retrieve a lexical match: they must identify relevant facts, recover dependencies among them, and compose a justified answer. Consider a bridge question that asks for an attribute of ``the director of film $F$.'' One passage identifies the director and a second supplies the requested attribute. If the question is written in Persian, the bridge passage in Japanese, and the answer-bearing passage in Spanish, the system must cross three knowledge interfaces: question-to-evidence access, transfer of the intermediate entity between evidence languages, and answer generation in the requested language. This setting tests whether a system can preserve a reasoning dependency while its linguistic realization changes at each role.

Figure~\ref{fig:worked-example} grounds this abstraction in an audited instance aligned with the worked example used in the source study. The question, the two gold paragraphs, and the answer occupy three languages, yet the annotated supporting sentences retain their paragraph and sentence identity.

\begin{figure}[htbp]
\centering
\begin{examplebox}{QUESTION --- Portuguese (pt)}{XHDeepBlue}
\emph{Qual s\'erie de fantasia cient\'ifica para jovens adultos, contada em primeira pessoa, possui um conjunto de livros complementares que narram as hist\'orias de mundos escravizados e esp\'ecies alien\'igenas?}
\end{examplebox}

\begin{examplebox}{GOLD PARAGRAPH 1 --- Swedish (sv) --- Hork-Bajir-kr\"onikorna}{XHGreen}
\textbf{[0]} The Hork-Bajir Chronicles \"ar den andra f\"oljeboken till ``Animorphs''-serien, skriven av K. A. Applegate.\par
\textbf{[1]} Med avseende p\r{a} kontinuiteten inom serien utspelar den sig f\"ore bok \#23, ``The Pretender'', \"aven om h\"andelserna i ber\"attelsen intr\"affar mellan tiden f\"or ``The Ellimist Chronicles'' och ``The Andalite Chronicles''.\par
\textbf{[2]} Boken introduceras av Tobias, som flyger till dalen med de fria Hork-Bajir, d\"ar Jara Hamee ber\"attar historien om hur Yeerks f\"orslavade Hork-Bajir, och hur Aldrea, en Andalite, och hennes f\"oljeslagare, Dak Hamee, en Hork-Bajir, f\"ors\"okte r\"adda sin v\"arld fr\r{a}n invasionen.
\end{examplebox}

\begin{examplebox}{GOLD PARAGRAPH 2 --- Dutch (nl) --- Animorphs}{XHGreen}
\textbf{[0]} Animorphs is een sciencefantasy-serie van young adult-boeken geschreven door Katherine Applegate en haar echtgenoot Michael Grant, die samen schrijven onder de naam K. A. Applegate, en gepubliceerd door Scholastic.\par
\textbf{[1]} Het wordt in de eerste persoon verteld, waarbij alle zes de hoofdpersonages om de beurt de boeken vertellen vanuit hun eigen perspectieven.
\end{examplebox}

\begin{examplebox}{ENGLISH REASONING GLOSS --- Answer language: Portuguese (pt)}{XHGold}
The Swedish evidence links the companion-book and enslaved-alien-world clues to \emph{Animorphs}; the Dutch evidence establishes its young-adult science-fantasy genre and first-person narration. \textbf{Answer: Animorphs.}
\end{examplebox}
\caption{A worked cross-lingual bridge instance. All five displayed sentences are gold supporting facts; their bridge/answer functions are interpretive descriptions for exposition, not additional role labels in the released annotation.}
\label{fig:worked-example}
\end{figure}
\FloatBarrier

HotpotQA made this reasoning structure observable by pairing multi-document questions with distractor paragraphs and sentence-level supporting facts \citep{yang2018hotpotqa}. Subsequent resources broadened the space of reasoning chains: 2WikiMultiHopQA records explicit evidence paths \citep{ho2020twiki}, and MuSiQue constructs connected question graphs intended to reduce single-hop shortcuts \citep{trivedi2022musique}. These resources established a useful evaluation principle: an answer score alone is insufficient when a model can exploit memorized facts, annotation artifacts, or one relevant passage. Evidence identification and answer composition should be measured jointly.

Multilingual question answering (QA) developed through a partly separate line of work. XQuAD and MLQA provide parallel extractive evaluation across languages \citep{artetxe2020xquad,lewis2020mlqa}; MKQA aligns language-independent answer representations across 26 languages \citep{longpre2021mkqa}; XOR QA tests retrieval when a non-English question may require English evidence \citep{asai2021xor}; and XRAG evaluates cross-lingual retrieval-augmented generation over multilingual documents \citep{liu2025xrag}. These resources test important forms of transfer and retrieval, but a whole-example translation does not isolate the language boundaries inside a reasoning chain. A model can perform well when question and evidence are translated together yet fail when the entity that closes hop~1 must be recognized in the language of hop~2.

Recent work confirms that this distinction matters. Pt-HotpotQA evaluates original and Portuguese-translated HotpotQA \citep{mucciaccia2025pthotpotqa}. Meng et al.\ construct a controlled set of 182 two-document questions across English and four target languages and report separable bridge, composition, and answer-hop failures \citep{meng2026language}. DaPT translates three multi-hop benchmarks into five languages and constructs dual source/English reasoning paths for multilingual RAG \citep{wang2026dapt}. These studies mean that a broad claim such as ``the first multilingual multi-hop benchmark'' is neither necessary nor defensible. The open resource question is more specific: can a large benchmark expose the language of each reasoning role, retain distractors and sentence-level evidence labels, and cover a sufficiently broad language inventory to support systematic diagnosis?

We introduce \dataset\ for that purpose. It transforms the HotpotQA supplied-candidate distractor task into a controlled environment for cross-lingual knowledge composition. The question and answer of an instance share one assigned language, whereas every candidate title and paragraph receives a language independently. Paragraph order, sentence order, answer supervision, question type, and supporting-fact indices remain explicit, so the benchmark can evaluate both evidence selection and answer composition. Our audit verifies the recovered positional join and identifies the structural conditions that a corrected canonical release must repair. The resulting task differs from both monolingual translation, where an entire instance uses one language, and generalized cross-lingual transfer, where one question language is paired with one context language. Here, linguistic state is a vector over reasoning roles and distractors, and the evidence graph makes those interfaces measurable.

This article studies five questions:

\begin{description}[leftmargin=1.0cm,style=nextline]
\item[RQ1] How much multi-hop competence transfers from direct prompting or English-only task adaptation when the evidence set is mixed-language?
\item[RQ2] Which module shows the larger observed within-task adaptation change: evidence selection or answer generation from supplied evidence?
\item[RQ3] How much performance is recovered by translate-train adaptation, and what additional effect is associated with teacher-generated rationales?
\item[RQ4] How are question--evidence alignment and script relation associated with reader and selector performance, and which claims remain identifiable without a paired intervention?
\item[RQ5] What does a balanced, source-referenced translation audit reveal about V1 and the supplied V2 release candidate, and which release claims remain unsupported?
\end{description}

The work makes four contributions to knowledge-based AI systems:

\begin{enumerate}[leftmargin=*,label=\textbf{C\arabic*.},labelsep=0.7em]
\item \textbf{Benchmark.} A controlled cross-lingual knowledge-composition benchmark with supplied distractors, sentence-level evidence supervision, 15,661 training instances, and 7,405 validation instances.
\item \textbf{Diagnostic framework.} A role-aware language geometry over the question, bridge evidence, answer-hop evidence, and distractors, together with measurable interface, script, diversity, and entropy descriptors.
\item \textbf{Empirical findings.} Modular selector--reader experiments show that cross-lingual composition is substantially more language-sensitive than supplied-candidate selection in the evaluated systems, while translate-train adaptation recovers much of the loss.
\item \textbf{Reproducibility.} An audited, versioned, provenance-aware release that preserves source alignment, prompt contracts, and explicit status boundaries for future benchmark revisions.
\end{enumerate}

The scientific claim is deliberately bounded. \dataset\ is not a naturally authored multilingual information-seeking corpus and it does not evaluate full-Wikipedia retrieval. It is an English-centric, translation-derived stress test for \emph{within-instance cross-lingual evidence selection and composition} under fixed candidate recall. This controlled scope makes language boundaries measurable, while the limitations make translation and cultural audits central to responsible use.

\section{Related work}\label{sec:related}

\subsection{Multi-hop question answering resources}

The general MHQA problem can be viewed as approximating a function from a question and multiple contexts to an answer, where at least one intermediate conclusion is a necessary premise for the next \citep{mavi2024survey}. HotpotQA operationalizes this idea with bridge and comparison questions, approximately ten candidate paragraphs in the distractor setting, and supporting sentences that make the evidence path partially observable \citep{yang2018hotpotqa}. 2WikiMultiHopQA adds structured evidence triples and evaluates answer, support, and evidence jointly \citep{ho2020twiki}. MuSiQue represents a composed question as a directed acyclic graph (DAG) of subquestions and explicitly tests whether predecessor answers are necessary \citep{trivedi2022musique}. These formulations motivate our distinction between merely seeing multiple documents and requiring a connected evidence chain.

Modeling work has addressed iterative retrieval, graph reasoning, question decomposition, and explicit reasoning chains. Chen et al.\ extract discrete chains of sentences before answer prediction \citep{chen2019chains}. Multi-step entity-centric retrieval uses entities recovered at one hop to retrieve the next \citep{godbole2019retrieval}. IRCoT alternates retrieval with chain-of-thought generation \citep{trivedi2023ircot}, while EfficientRAG and generate-then-ground systems improve the efficiency or grounding of multi-hop retrieval \citep{zhuang2024efficientrag,shi2024gtg}. \dataset\ is complementary: it does not propose a new retrieval architecture, but makes the language of every supplied context an explicit experimental variable.

\subsection{Multilingual and cross-lingual QA}

Parallel benchmarks provide controlled transfer comparisons. XQuAD translates a subset of SQuAD into ten languages \citep{artetxe2020xquad}; MLQA supplies multi-way aligned contexts, questions, and answers in seven languages and defines generalized transfer where question and context languages differ \citep{lewis2020mlqa}. MKQA offers 10,000 questions with aligned answers across 26 languages \citep{longpre2021mkqa}. XOR QA instead begins with naturally authored questions in seven non-English languages and permits retrieval from English and multilingual collections \citep{asai2021xor}. Mintaka includes complex multilingual questions grounded in structured knowledge \citep{sen2022mintaka}, and JEMHopQA contributes explainable Japanese multi-hop QA \citep{ishii2024jemhopqa}.

These resources differ along three axes that are sometimes collapsed under ``cross-lingual QA'': the language in which a model was trained, the language interface between question and evidence, and the language composition within the evidence itself. We use \emph{transfer} for the first, \emph{interface crossing} for the second, and \emph{within-chain crossing} for the third. This terminology prevents a model trained on one language and tested on another from being treated as equivalent to a model that must combine two simultaneously visible evidence languages.

\subsection{Cross-lingual multi-hop and RAG benchmarks}

Pt-HotpotQA studies a complete Portuguese translation of HotpotQA and therefore provides an important same-language multilingual control \citep{mucciaccia2025pthotpotqa}. XRAG combines questions and documents across languages in a retrieval-augmented generation setting, with two supporting and six distractor documents \citep{liu2025xrag}. Meng et al.\ isolate a path $q\rightarrow D_1\rightarrow b\rightarrow D_2\rightarrow a$ and change the languages of the two gold documents \citep{meng2026language}. Their filtered 182-item study is particularly close to our role-level analysis. DaPT translates subsets of HotpotQA, 2WikiMultiHopQA, and MuSiQue into five languages and constructs source- and English-path subquestion graphs \citep{wang2026dapt}.

Relative to this literature, \dataset\ combines 24-language coverage, supplied distractor candidates, all 7,405 HotpotQA distractor validation questions, and sentence-level supporting-fact labels. It is larger and linguistically broader than the controlled probe of Meng et al., but less ecologically realistic than naturally authored questions or open-corpus multilingual retrieval. Table~\ref{tab:resource-comparison} states this positioning without a priority claim.

\begin{table}[htbp]
\caption{Positioning of representative multilingual and cross-lingual QA resources. ``Mixed evidence'' describes the language boundary visible within an input, rather than only the language in which a model is trained.}
\label{tab:resource-comparison}
\footnotesize
\begin{tabularx}{\linewidth}{@{}lrrlYYY@{}}
\toprule
Resource & Lang. & Items & MHQA & Mixed evidence & Evidence label & Scope \\
\midrule
MLQA & 7 & 12k & No & question--context pairs & answer spans & fixed context \\
XOR QA & 7+En & 40k & not primary & non-English query, multilingual corpus & answer evidence & open retrieval \\
Pt-HotpotQA & 2 & 113k & Yes & whole-example translation & sentences & fixed/full wiki \\
XRAG & 5 & 4k & cross-document & cross-lingual RAG documents & documents & fixed candidates \\
Meng et al. & 5 & 182 & two-hop & separate gold-hop languages & documents & oracle documents \\
DaPT & 5 & 3k & Yes & source--English paths & dataset-dependent & RAG \\
\dataset & 24 & 23,066$^{b}$ & Yes & independent candidate languages$^{a}$ & sentences & supplied candidates \\
\bottomrule
\end{tabularx}
\begin{flushleft}\scriptsize $^{a}$The schema records bridge and answer-hop roles when determinable; assignment remains paragraph-level and preserves source support labels. $^{b}$The public audited V1.1 base has 15,661 training and 7,405 validation rows and includes original English source sentences; the recovered archive additionally retains 375,864 parallel training views.\end{flushleft}
\end{table}

\subsection{Translation-derived resources and evaluation validity}

Translation enables broad coverage at a fraction of natural data-collection cost, but it changes the object being measured. Source-language cultural content persists; translationese can simplify or distort lexical relations; entities may be transliterated differently across question and evidence; and answer aliases may not survive exact-match normalization. Thellmann et al.\ quantify a relationship between translation errors and downstream multilingual benchmark scores \citep{thellmann2026translation}. Data statements and datasheets consequently recommend documenting source populations, collection, transformations, intended uses, and limitations \citep{bender2018datastatements,gebru2021datasheets}. We treat translation quality and label preservation as parts of benchmark validity, not as incidental preprocessing.

A companion preprint, Bactrainus, studies a modular selector--reader architecture and evidence control on English HotpotQA \citep{barati2025bactrainus}. The present contribution is distinct: it formalizes mixed-language knowledge composition, evaluates language roles and interfaces, and treats resource validity as part of benchmark design. English-tuned modules appear only as transfer baselines.

\section{Formal problem definition}\label{sec:formal}

\subsection{Instances, facts, and evidence chains}

Let $\mathcal L$ be a finite language inventory and let $\mathbb C$ be a universe of contexts. A fixed-candidate MHQA instance is
\begin{equation}
x=(q,a,\mathcal C,\mathcal S^{\star},\mathcal G,\lambda),
\label{eq:instance}
\end{equation}
where $q$ is a question, $a$ is a gold answer, $\mathcal C=(c_1,\ldots,c_m)$ is an ordered candidate sequence, $\mathcal S^{\star}$ is the gold supporting-fact set, $\mathcal G$ is an evidence-dependency graph, and $\lambda$ maps linguistic units to languages. Candidate
\begin{equation}
c_i=(u_i,s_{i1},\ldots,s_{in_i})
\end{equation}
contains a title $u_i$ and ordered sentences. Its atomic fact identifiers are pairs $(i,j)$, so
\begin{equation}
\mathcal F(\mathcal C)=\{(i,j):1\le i\le m,\ 1\le j\le n_i\},
\qquad \mathcal S^{\star}\subseteq\mathcal F(\mathcal C).
\label{eq:facts}
\end{equation}
This representation separates a stable annotation key $(i,j)$ from the translated surface form $s_{ij}$. It is essential for resource integrity: a translated title cannot silently change the identity of a supporting fact.

The broad definition of MHQA approximates a function $f:\mathcal Q\times\mathbb C^m\rightarrow\mathcal A\cup\{\Phi\}$ over multiple contexts \citep{mavi2024survey}. We make the notion of a hop explicit with an abstract inference relation $\vdash$. A $k$-hop evidence chain is organized into inference-stage evidence blocks $E_1,\ldots,E_k$, where each non-empty $E_h\subseteq\mathcal S^{\star}$ may contain more than one annotated sentence:
\begin{equation}
\pi=(E_1,\ldots,E_k;z_1,\ldots,z_k),\qquad
\bigcup_{h=1}^{k}E_h\subseteq\mathcal S^{\star},
\label{eq:chain}
\end{equation}
such that
\begin{align}
(q,E_1)&\vdash z_1,\\
(q,z_{1:h-1},E_h)&\vdash z_h \quad (2\le h\le k),\\
z_k&\models a.
\label{eq:chain-inference}
\end{align}
The intermediate $z_h$ may be an entity, relation, numerical result, or proposition. The evidence graph $\mathcal G=(V,E)$ contains nodes for $q$, evidence facts, intermediate conclusions, and $a$; a directed edge denotes premise dependence. This follows the connected-question view of MuSiQue while keeping the context granularity of HotpotQA \citep{trivedi2022musique}.

Merely placing several facts in a prompt does not prove multi-hop necessity. We therefore distinguish an annotated chain from a necessary chain. Let a model-independent semantic oracle $\Omega$ return one exactly when a proposed inference chain is valid. Define
\begin{equation}
k^{\star}_{\Omega}(x)=\min\{k:\exists(E_1,\ldots,E_k;z_1,\ldots,z_k),
\ \Omega(q,E_{1:k},z_{1:k},a)=1\},
\label{eq:minimal-hop}
\end{equation}
where oracle acceptance requires Eqs.~\ref{eq:chain}--\ref{eq:chain-inference}. This counts dependency stages rather than annotated sentences. The instance is strictly multi-hop under this abstraction when $k^{\star}_{\Omega}(x)\ge2$ and no proper predecessor-closed subset of the gold evidence graph entails $a$. In practice, parametric model knowledge makes necessity model-dependent. Partial-evidence tests therefore estimate
\begin{equation}
N_h(M,x)=\ind[M(q,\mathcal S^{\star})=a]-
\ind[M(q,\mathcal S^{\star}\setminus E_h)=a],
\label{eq:necessity}
\end{equation}
rather than claiming a complete logical proof.

\begin{table}[htbp]
\caption{Core notation used throughout the article.}
\label{tab:notation}
\small
\begin{tabularx}{\linewidth}{@{}lY@{}}
\toprule
Symbol & Meaning \\
\midrule
$x$ & one benchmark instance (Eq.~\ref{eq:instance}) \\
$q,a$ & question and gold answer \\
$\mathcal C=(c_1,\ldots,c_m)$ & ordered candidate paragraphs \\
$\mathcal S^{\star},\widehat{\mathcal S}$ & gold and predicted supporting facts \\
$\mathcal G$ & evidence-dependency graph \\
$\lambda(z)$ & language assigned to unit $z$ \\
$\ell_q,\ell_a,\ell_b,\ell_t$ & question, answer, bridge-hop, and answer-hop languages \\
$\rho_G,\rho_D$ & question mismatch for gold and distractor paragraphs; $\rho_D$ may be undefined \\
$H_G,H_C$ & normalized entropy of gold and all-candidate languages \\
$K_C$ & number of distinct languages among candidate paragraphs \\
$M(x)$ & an instance-level answer, support, or joint metric \\
\bottomrule
\end{tabularx}
\end{table}

\subsection{Reader, selector, and end-to-end tasks}

A full system predicts both evidence and answer,
\begin{equation}
f_{\theta}(q,\mathcal C,\lambda)=(\widehat a,\widehat{\mathcal S}).
\label{eq:system}
\end{equation}
The modular baseline factorizes the conditional distribution as
\begin{equation}
p_{\theta}(\widehat a,\widehat{\mathcal S}\mid q,\mathcal C,\lambda)
=p_{\theta_s}(\widehat{\mathcal S}\mid q,\mathcal C,\lambda)
\,p_{\theta_r}(\widehat a\mid q,\widehat{\mathcal S},\lambda).
\label{eq:factorization}
\end{equation}
This yields three nested tasks. The historical \emph{reader} receives $q$ and the annotated supporting sentences $\{s_{ij}:(i,j)\in\mathcal S^{\star}\}$, grouped under their source-paragraph titles, and predicts $a$; it therefore isolates answer composition from evidence selection without exposing the non-supporting sentences of a gold paragraph. The \emph{selector} receives all $m$ supplied candidates and predicts $\mathcal S^{\star}$, isolating selection from corpus retrieval. The \emph{end-to-end} task evaluates Eq.~\ref{eq:system}. We use ``selection'' rather than ``retrieval'' because candidate recall is fixed by the HotpotQA distractor input.

\subsection{Language map and cross-lingual conditions}

For any linguistic unit $z$, $\lambda(z)\in\mathcal L$. \dataset\ imposes
\begin{equation}
\lambda(q)=\lambda(a)=\ell_q,\qquad
\lambda(c_i)\sim\operatorname{Unif}(\mathcal L)
\quad\text{independently for }i=1,\ldots,m.
\label{eq:assignment}
\end{equation}
Titles and sentences inside one paragraph share $\lambda(c_i)$. For a bridge question with two support roles, define the language-state vector
\begin{equation}
\vec{\ell}_x=(\ell_q,\ell_a,\ell_b,\ell_t,
\ell_{d_1},\ldots,\ell_{d_{m-g}}),
\label{eq:language-vector}
\end{equation}
where $b$ resolves an intermediate bridge and $t$ contains the answer-bearing fact. Comparison questions have an unordered pair of support roles rather than a unique bridge/target order.

Let the unique gold-paragraph set and its complement be
\begin{equation}
\mathcal C_G(x)=\{c_i\in\mathcal C:\exists j,(i,j)\in\mathcal S^{\star}\},
\qquad \mathcal C_D(x)=\mathcal C\setminus\mathcal C_G(x).
\label{eq:gold-paragraph-set}
\end{equation}
The map $\lambda(c_i)$ denotes the assigned target-language label of the entire candidate, not a post-hoc language-ID prediction. For marginal analysis, define
\begin{equation}
G_{\ell}(x)=\ind\!\left[\exists c\in\mathcal C_G(x):\lambda(c)=\ell\right].
\label{eq:gold-language-membership}
\end{equation}
Thus each source contributes at most once to the marginal row for language $\ell$, even if it contains several supporting sentences in that assigned language.

Two forms of crossing follow. \emph{Interface crossing} occurs when at least one gold evidence paragraph differs from the question language:
\begin{equation}
I(x)=\ind\!\left[\exists c\in\mathcal C_G:\lambda(c)\ne\ell_q\right].
\label{eq:interface}
\end{equation}
\emph{Composition crossing} occurs when the gold evidence itself is multilingual:
\begin{equation}
C(x)=\ind\!\left[\left|\{\lambda(c):c\in\mathcal C_G\}\right|>1\right].
\label{eq:composition}
\end{equation}
Thus $I(x)=1,C(x)=0$ describes a question/evidence interface change with monolingual evidence, whereas $C(x)=1$ exposes a language boundary within the annotated gold evidence. It does not by itself prove that both paragraphs are necessary for a particular model.

We quantify mismatch over the unique paragraph sets in Eq.~\ref{eq:gold-paragraph-set} as
\begin{align}
\rho_G(x)&=\frac{1}{|\mathcal C_G|}\sum_{c\in\mathcal C_G}
\ind[\lambda(c)\ne\ell_q],\\
\rho_D(x)&=
\begin{cases}
\displaystyle\frac{1}{|\mathcal C_D|}\sum_{c\in\mathcal C_D}
\ind[\lambda(c)\ne\ell_q],&|\mathcal C_D|>0,\\[5pt]
\mathrm{NA},&|\mathcal C_D|=0.
\end{cases}
\label{eq:mismatch}
\end{align}
For $R\in\{G,C\}$, let $p_{R,\ell}$ be the fraction of paragraphs in $\mathcal C_R$ assigned language $\ell$, with $\mathcal C_C=\mathcal C$. Normalized evidence-language entropy and candidate-language richness are
\begin{equation}
H_R(x)=
\begin{cases}
-\displaystyle\frac{\sum_{\ell\in\mathcal L}p_{R,\ell}\log p_{R,\ell}}
{\log |\mathcal C_R|},& |\mathcal C_R|>1,\\[4pt]
0,&|\mathcal C_R|\le1,
\end{cases}
\qquad K_C(x)=\left|\{\lambda(c):c\in\mathcal C\}\right|.
\label{eq:entropy}
\end{equation}
Paragraph-level weighting prevents a paragraph with many annotated sentences from dominating the linguistic descriptor.

Equations~\ref{eq:mismatch}--\ref{eq:entropy} define five descriptive strata for the observed validation geometry: (S0) $\rho_G=\rho_D=0$; (S1) $\rho_G=0,\rho_D>0$; (S2) $0<\rho_G<1$; (S3) $\rho_G=1$ with one gold language; and (S4) $\rho_G=1$ with multiple gold languages. S0 and S1 require $\rho_D$ to be defined; a no-distractor item with $\rho_G=0$ would receive a separate NA descriptor rather than being coded as S0. Analyses involving $\rho_D$ omit no-distractor items rather than coding missingness as zero. No stratum is assumed to be populated. In particular, with ten independently assigned candidates, $\Pr(\mathrm{S0})=24^{-10}$, so S0 is not a usable empirical reference group in a 7,405-item sample. Role-aware analyses additionally use the equalities among $\ell_q,\ell_b,\ell_t$ in Table~\ref{tab:role-conditions}.

\begin{table}[htbp]
\caption{Role-aware conditions for two-hop bridge questions.}
\label{tab:role-conditions}
\small
\begin{tabularx}{\linewidth}{@{}l l Y@{}}
\toprule
Condition & Constraint & Primary capability \\
\midrule
Monolingual & $\ell_q=\ell_b=\ell_t$ & reasoning without language switching \\
Query mismatch & $\ell_q\ne\ell_b=\ell_t$ & question-to-evidence access \\
Bridge mismatch & $\ell_q=\ell_t\ne\ell_b$ & intermediate entity transfer \\
Answer-hop mismatch & $\ell_q=\ell_b\ne\ell_t$ & answer recovery across language \\
Cross-hop & $\ell_b\ne\ell_t$ & within-chain composition \\
Three-language & $|\{\ell_q,\ell_b,\ell_t\}|=3$ & maximal role switching \\
\bottomrule
\end{tabularx}
\end{table}

\subsection{Evaluation measures}

Let $n_{\ell}$ be a language-aware normalization function and $T_{\ell}$ its tokenization. Answer exact match for acceptable answer set $A_x$ is
\begin{equation}
\operatorname{EM}_a(x)=\max_{a'\in A_x}
\ind[n_{\ell_q}(\widehat a)=n_{\ell_q}(a')].
\label{eq:answer-em}
\end{equation}
The present schema supplies one canonical answer, hence $A_x=\{a\}$ for the current evaluator; a future version may add adjudicated aliases without changing the definition.
Answer precision and recall use multiset token overlap $o=\sum_t\min(C_{\widehat a}(t),C_{a'}(t))$:
\begin{equation}
P_a=\frac{o}{|T_{\ell_q}(\widehat a)|},\quad
R_a=\frac{o}{|T_{\ell_q}(a')|},\quad
F_{1,a}=\frac{2P_aR_a}{P_a+R_a}.
\label{eq:answer-f1}
\end{equation}
For predicted and gold fact sets,
\begin{equation}
P_s=\frac{|\widehat{\mathcal S}\cap\mathcal S^{\star}|}{|\widehat{\mathcal S}|},\quad
R_s=\frac{|\widehat{\mathcal S}\cap\mathcal S^{\star}|}{|\mathcal S^{\star}|},\quad
F_{1,s}=\frac{2P_sR_s}{P_s+R_s},\quad
\operatorname{EM}_s=\ind[\widehat{\mathcal S}=\mathcal S^{\star}].
\label{eq:support-f1}
\end{equation}
The evaluator assigns the usual zero-safe values when a predicted or gold token/fact set is empty.
Hotpot-style joint metrics multiply component precision and recall before taking the harmonic mean,
\begin{equation}
P_j=P_aP_s,\quad R_j=R_aR_s,\quad
F_{1,j}=\frac{2P_jR_j}{P_j+R_j},\quad
\operatorname{EM}_j=\operatorname{EM}_a\operatorname{EM}_s.
\label{eq:joint}
\end{equation}
Joint F1 is therefore not the product of answer F1 and support F1. We report instance-average metrics for HotpotQA compatibility and recommend macro-by-question-language aggregation,
\begin{equation}
\operatorname{MacroLang}(M)=\frac{1}{|\mathcal L|}
\sum_{\ell\in\mathcal L}\frac{1}{N_{\ell}}
\sum_{x:\ell_q=\ell}M(x),
\label{eq:macro}
\end{equation}
to prevent frequent language conditions from dominating the benchmark.

\section{Resource construction}\label{sec:construction}

\subsection{Source configuration and splits}

\dataset\ derives from HotpotQA's distractor configuration \citep{yang2018hotpotqa}. Training retains the 15,661 examples labeled \texttt{hard}; validation contains all 7,405 official distractor-development instances. The transformation does not create a hidden test set. Each item has approximately ten supplied paragraphs, normally two gold paragraphs plus distractors. The resource contains 23,066 base instances, 55,830 supporting-fact annotation occurrences, and 229,688 candidate-paragraph occurrences (Table~\ref{tab:sizes}).

\begin{table}[htbp]
\caption{\dataset\ split sizes. Support counts are annotation occurrences and document counts are candidate occurrences.}
\label{tab:sizes}
\centering
\begin{tabular}{@{}lrrr@{}}
\toprule
Split & Questions & Support facts & Candidate docs \\
\midrule
Train (hard) & 15,661 & 37,825 & 155,988 \\
Validation & 7,405 & 18,005 & 73,700 \\
\midrule
Total & 23,066 & 55,830 & 229,688 \\
\bottomrule
\end{tabular}
\end{table}

The source bridge/comparison labels and difficulty metadata are retained. Supporting facts remain anchored to immutable paragraph and sentence identifiers. Since candidates are supplied, reported selector results are conditional on candidate recall and must not be interpreted as full-Wikipedia retrieval.

\subsection{Language inventory}

The inventory is English, Mandarin Chinese, Hindi, Spanish, Arabic, French, Bengali, Portuguese, Russian, Urdu, Indonesian, German, Japanese, Turkish, Vietnamese, Swahili, Korean, Persian, Italian, Thai, Dutch, Polish, Greek, and Swedish. It spans Latin, Arabic, Devanagari, Bengali, Han, Japanese, Hangul, Greek, Cyrillic, and Thai scripts and several language families. Selection considered speaker population, web presence, contemporary model coverage, and typological/script variety. These are not all low-resource languages, and resource-level labels are not inferred from language identity.

Question-language assignment is close to uniform. Validation contains between 274 and 339 questions per language (mean 308.54, population SD 16.26, coefficient of variation 5.27\%). Candidate and support language counts are reported in Supplementary Section~A. Uniformity of marginal counts does not guarantee balance over ordered language pairs, scripts, or hop roles; the descriptors in Section~\ref{sec:formal} are therefore necessary.

\subsection{Translation and assignment pipeline}

For each source instance, the intended transformation is:

\begin{enumerate}[leftmargin=*]
\item sample one language $\ell_q$ and translate the question and answer into that language;
\item independently sample a language for every candidate paragraph;
\item translate each title and its ordered sentence array while preserving paragraph identity, sentence cardinality, and order;
\item attach the original support indices, question type, difficulty, language codes, and transformation provenance; and
\item in the canonicalizer, reject or quarantine records that violate format, declared-language, alignment, or record-integrity invariants.
\end{enumerate}

The audited legacy generators did not execute these steps symmetrically. For every hard training source they serialized all 24 question--answer translations consecutively while keeping one mixed-language context fixed, producing $15{,}661\times24=375{,}864$ rows. A later notebook selected one row from each 24-row group with an unseeded pseudorandom draw to form the historical 15,661-row base train split. Validation instead serialized one randomly selected question--answer language per source, producing 7,405 rows. The supplied shards have contiguous indices and exact source order, but contain no source IDs; we recover those IDs by an audited positional join to the official HotpotQA order. The public audited V1.1 configuration materializes the recovered IDs, positional provenance, and original English source sentences so consumers do not depend on position or a separate source lookup.

\begin{figure}[htbp]
\centering
\includegraphics[width=0.98\linewidth]{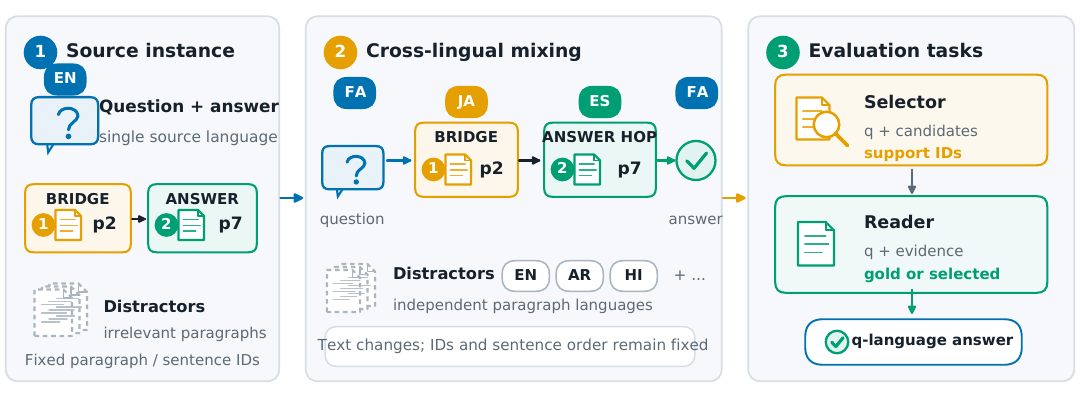}
\caption{\dataset\ transformation and task decomposition. Stable paragraph/sentence IDs preserve supervision while the question--answer pair and candidate paragraphs receive languages independently. Bridge and answer-hop roles define within-chain crossing; distractors define irrelevant-language interference.}
\label{fig:design}
\end{figure}

Figure~\ref{fig:design} illustrates the result. The selector observes all mixed-language candidates and predicts evidence; the reader returns an answer in $\ell_q$. Because title and body move together, the design does not create an artificial title/body language boundary.

\FloatBarrier

The evaluated V1 resource used these historical random assignments. Its validation mapping and 24-view training groups are recoverable, but the unseeded base-train draw is not. Public V1.1 therefore freezes and labels a reproducible one-view projection while retaining the English source sentences; it does not claim to reconstruct the lost draw. V2 uses a separately realized assignment, so its translation-quality audit is independent rather than paired. Supplementary Section~C documents the version-specific release checks.

\subsection{Translator selection and model freeze}

The translator pilot compared GPT-4o, GPT-4o mini, and Llama~3.1~70B Instruct. Ten Persian examples were assessed by three specialists; a follow-up used two examples for each of the 23 translated languages with GPT-4o-based comparative judging. GPT-4o mini was selected because its quality/cost trade-off was preferred for corpus-scale translation. The pilot motivates the choice but is too small to estimate final error rates for every language.

At the January 2025 experimental cutoff, GPT-4o and GPT-4o mini were the latest GPT variants considered for reproduction \citep{openai2024gpt4o}. The historical V1 generator used GPT-4o mini and constrained each response to the requested translation. The V2 release candidate was generated with Gemma~4~31B Instruct \citep{gemma4report}; it preserves the exact English source sentences beside the translated evidence. Supplementary Section~F reproduces the translation prompts used by both versions.

\subsection{Expected language mixing under the design}

Independent assignment makes the severity of language mixing analytically transparent. With $L=|\mathcal L|$ and $g$ distinct gold paragraphs,
\begin{align}
\Pr[I(x)=1]&=1-L^{-g},\\
\Pr[C(x)=1]&=1-L^{1-g}.
\label{eq:mix-prob}
\end{align}
For the typical $L=24,g=2$ case, at least one gold paragraph differs from the question with probability $99.826\%$, and the two gold paragraphs differ from each other with probability $95.833\%$. Conditioned on an observed candidate count $m_i$, the question language is absent from the candidate set with probability
\begin{equation}
\Pr[\ell_q\notin\lambda(\mathcal C_i)\mid m_i]
=\left(1-\frac{1}{L}\right)^{m_i},
\label{eq:q-absent}
\end{equation}
and the expected number of distinct candidate languages is
\begin{equation}
\mathbb E[K_C(x_i)\mid m_i]
=L\left[1-\left(1-\frac{1}{L}\right)^{m_i}\right].
\label{eq:distinct-languages}
\end{equation}
The validation audit finds $m_i=10$ for 7,345 items and $m_i\in[2,9]$ for the remaining 60. Averaging Eqs.~\ref{eq:q-absent}--\ref{eq:distinct-languages} over those observed counts gives 65.490\% expected question-language absence and $\mathbb E[K_C]=8.282$. The realized values are 65.523\% and 8.300. Likewise, $I(x)=1$ in 7,391 items (99.811\%; IID expectation 99.826\%) and $C(x)=1$ in 7,079 (95.598\%; expectation 95.833\%). Thus the intended high-mixing regime is realized without an anomalous assignment imbalance. The S0--S4 counts are $0/14/564/312/6{,}515$; S1 is too small for inference.

\begin{figure}[htbp]
\centering
\includegraphics[width=0.98\linewidth]{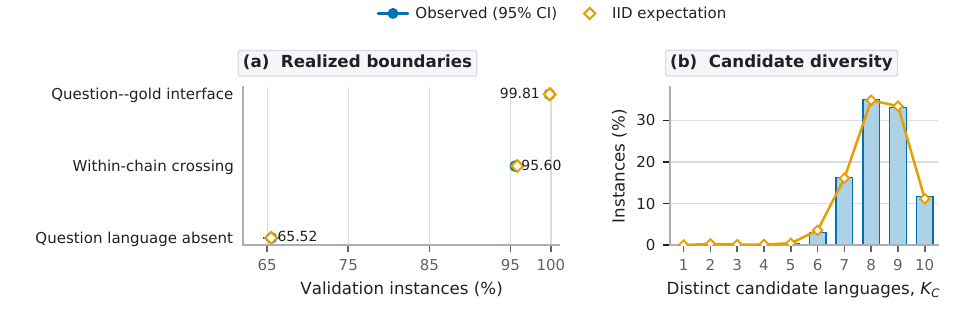}
\caption{Realized validation mixing versus the exact IID assignment model. (a) Observed rates and Wilson 95\% intervals; IID expectations condition on two gold paragraphs and each item's actual candidate count. (b) Observed and expected candidate-language occupancy. Mean $K_C$ is 8.300 observed versus 8.282 expected ($n=7{,}405$).}
\label{fig:mixing}
\end{figure}
\FloatBarrier

\subsection{Parallel views and the status of \datasetplus}

The \datasetplus\ protocol holds the mixed-language candidate set fixed and pairs it with all 24 translations of the question and answer. If complete for both base splits, it contains
\begin{equation}
15{,}661\times24=375{,}864\ \text{training views},\qquad
7{,}405\times24=177{,}720\ \text{validation views},
\end{equation}
or 553,584 views in total. The audited archive contains the complete 375,864-row training half: every source has the exact 24-language order and byte-identical context across its views. It does \emph{not} contain the missing 23 question--answer translations per validation source, so the 177,720-row validation half and hence the complete 553,584-view resource remain a prospective \datasetplus\ generation-and-release target, not a current release claim. The paired design supports question-language effects because evidence content and placement remain fixed; uncertainty must cluster by $\texttt{source\_id}$ rather than treating repeated views as independent.

\subsection{Release status}

The public V1.1 configuration retains all 15,661 training and 7,405 validation instances, explicit audit status, and the original English source sentences. The Gemma-generated V2 candidate contains 22,836 of the intended 23,066 instances and is therefore released as an \emph{audited incomplete snapshot}, not as a replacement for V1. Both absent validation sources violate upstream HotpotQA invariants: one supplied paragraph contains a blank-string sentence, and one supporting-fact annotation uses sentence index 902 outside the paragraph bounds. These inherited defects triggered the omissions, but omission is the current V2 workflow's treatment of them rather than a property of the source rows alone. Across both splits, 50 of the 230 absent rows are source-triggered and the remaining 180 are clean-source training-generation omissions; Supplementary Section~C gives the complete attribution. A completed revision requires source-ID coverage, explicit treatment of inherited annotations, and bilingual review; the parallel validation portion of \datasetplus\ also remains future work.

\section{Experimental protocol}\label{sec:experiments}

\subsection{Evaluation tasks}

The historical experiment tables evaluate the 7,405-instance validation split. The reader receives annotated supporting sentences grouped under their source-paragraph titles; the selector receives all supplied candidates and predicts sentence-level support; the end-to-end condition passes selected evidence to the reader. This separation maps directly to Eq.~\ref{eq:factorization}: reader results measure composition under oracle evidence, whereas selector and joint results expose the evidence bottleneck. Complete prediction artifacts survive for three one-shot readers and the adapted selector; absent responses are scored as empty. Other values are retained as historical aggregate results.

\subsection{One-shot model screen}

The one-shot reader screen covers proprietary and open multilingual instruction models from 7B to 405B parameters. Each model receives one demonstration, a question, and the annotated supporting sentences grouped under source titles; it must return only an answer in the question language. The open families are Llama~3.1 Instruct \citep{grattafiori2024llama3}, Qwen2 Instruct \citep{yang2024qwen2}, Aya~23 \citep{aryabumi2024aya23}, and Gemma~2 Instruct \citep{gemmateam2024gemma2}. Supplementary Section~F tabulates the evaluated repository identifiers and reproduces the shared reader prompt. Because immutable serving records are unavailable for every historical endpoint, this screen is treated as a comparative baseline rather than a bitwise reproduction claim.

\subsection{Transfer and translate-train systems}

Four adaptation conditions are compared:

\begin{description}[leftmargin=1.4cm,style=nextline]
\item[Direct] Llama~3.1~8B Instruct with one demonstration and no task training.
\item[English transfer] the 8B Bactrainus reader or selector tuned on English HotpotQA and applied directly to \dataset.
\item[Translate-train] \xreader\ and \xselector, each initialized from Llama~3.1~8B Instruct and adapted on the 15,661 mixed-language training instances.
\item[Rationale continuation] \xreader\ receives an additional continuation stage with English rationales generated by a Llama~3.1~70B teacher. Since this stage changes supervision as well as training duration, its difference is treated as an associated gain rather than a pure estimate of rationale quality.
\end{description}

Both multilingual modules use LoRA \citep{hu2022lora}. Table~\ref{tab:training} summarizes the common recipe. The selector's longer sequence limit reflects the full candidate set; the reader operates on gold or selected evidence.

\begin{table}[htbp]
\caption{Translate-train configuration for the two Llama~3.1~8B modules. QKVO denotes attention projections; MLP denotes feed-forward projections.}
\label{tab:training}
\small
\begin{tabularx}{\linewidth}{@{}lrrY@{}}
\toprule
Setting & Reader & Selector & Shared value \\
\midrule
Training records & 15,661 & 15,661 & 2 reported epochs \\
Per-device batch & 8 & 2 & -- \\
Gradient accumulation & 32 & 16 & -- \\
Maximum length & 512 & 4,096 & -- \\
LR schedule & \multicolumn{2}{c}{cosine decay} & warm-up 0.03 \\
Peak learning rate & \multicolumn{2}{c}{$10^{-4}$} & dropout 0.05 \\
LoRA & \multicolumn{2}{c}{$r=64,\alpha=128$} & QKVO, MLP, LM head \\
\bottomrule
\end{tabularx}
\end{table}

\subsection{Normalization and aggregation}

The primary answer metrics follow Eqs.~\ref{eq:answer-em}--\ref{eq:answer-f1}; support and joint metrics follow Eqs.~\ref{eq:support-f1}--\ref{eq:joint}. The current toolkit evaluator applies Unicode NFKC, case folding where defined, punctuation/whitespace normalization, and English article removal only for English. For Chinese, Japanese, and Thai, answer F1 uses character tokens rather than whitespace tokens. Macro-language, mismatch, entropy, candidate-diversity, and script aggregates are emitted by the toolkit. Output-language correctness requires a separately versioned language-identification or human-audit result and is not inferred from answer F1. The historical baseline tables below retain the original Hotpot normalization for comparability and are therefore labeled as legacy-metric results.

Historical transfer values are single-run percentages. For the 7,405-row prediction artifacts used in RQ4, we publish per-item scores and percentile bootstrap intervals; these quantify sampling uncertainty for one frozen run, not training-seed variation. Trained-model comparisons still require multiple seeds. For repeated \datasetplus\ views, resampling must be clustered by source ID.

\subsection{Source-referenced LLM-as-judge audit}\label{sec:judge-method}

We compare translation units with a frozen source-referenced LLM-as-judge protocol. For each of the 23 non-English target languages and each version, a seeded sampler draws 80 paragraph translations, 20 questions, and 20 answers from the combined training and validation pools: $120$ units per language and $2{,}760$ per version. The V1 and V2 samplers operate on different realized resources, so their samples are independent rather than matched; exact unit overlap is negligible and no paired test is used. The sampling seed is 20260810.

The requested judge alias was GLM-5.2 \citep{glm5team2026glm5,zai2026glm52}. Each unit receives one source-referenced integer score in $[0,100]$. Paragraph and question rubrics emphasize adequacy, entity and number preservation, fluency, and style; the answer rubric additionally emphasizes factual identity and polarity. This model-based audit supports triage and comparison, not bilingual certification. A misconfigured preliminary answer subset was excluded before analysis and replaced by a dedicated answer run, yielding 2,760 valid units for each version. Supplementary Section~D reports the construction, distributions, language-level results, and exact judge prompts.

For each unit type and the pooled mean, we report a 20,000-replicate percentile bootstrap that resamples V1 and V2 independently within language-by-unit strata (analysis seed 20260821). Let $\bar s_{vu}^{*(b)}$ be bootstrap replicate $b$ for version $v$ and unit $u$. The descriptive difference interval is the percentile interval of
\begin{equation}
D_u^{*(b)}=\bar s_{2u}^{*(b)}-\bar s_{1u}^{*(b)},
\label{eq:judge-bootstrap}
\end{equation}
where the two terms are generated by independent resamples. The pooled estimator preserves the fixed 80:20:20 paragraph/question/answer weighting. These intervals quantify sampling variation under the audit design; they do not include judge-model, prompt, or endpoint uncertainty.

\section{Results}\label{sec:results}

\subsection{One-shot multilingual reader capability}

Table~\ref{tab:screen} shows a broad performance range. GPT-4o reaches the highest one-shot reader F1 (50.82), followed by Llama~3.1~405B (48.32) and GPT-4o mini (45.13). Larger models are generally stronger within a family, but parameter count is not a sufficient explanation across families: Aya~23~35B (36.92) trails Gemma~2~27B (39.54), while the artifact historically labeled Llama~3.1~70B (42.80) trails the lower-cost GPT-4o mini endpoint. Multilingual pretraining, instruction tuning, output-language control, and answer normalization may contribute.

\begin{table}[htbp]
\caption{Historically reported one-shot screening aggregates with gold supporting facts on $n=7{,}405$ under legacy Hotpot normalization. $\dagger$ marks systems for which a 7,405-row local artifact permits the independent re-score in Supplementary Section~B. For the 70B Llama row, the model name is a historical artifact label whose served checkpoint awaits server-log verification.}
\label{tab:screen}
\centering
\small
\begin{tabular}{@{}lrr@{}}
\toprule
Model & Answer EM & Answer F1 \\
\midrule
GPT-4o & \textbf{35.10} & \textbf{50.82} \\
GPT-4o mini$^{\dagger}$ & 29.94 & 45.13 \\
Llama 3.1 Instruct 405B & 33.76 & 48.32 \\
Qwen2 Instruct 72B$^{\dagger}$ & 24.94 & 37.95 \\
Llama 3.1 Instruct 70B$^{\dagger}$ & 30.74 & 42.80 \\
Aya 23 35B & 25.43 & 36.92 \\
Gemma 2 27B & 28.51 & 39.54 \\
Gemma 2 9B & 24.83 & 35.32 \\
Llama 3.1 Instruct 8B (screening run) & 21.62 & 32.12 \\
Aya 23 8B & 15.32 & 22.29 \\
Qwen2 Instruct 7B & 19.67 & 28.12 \\
\bottomrule
\end{tabular}
\end{table}

The best one-shot F1 remains below 51 despite oracle supporting sentences. This is important: failures are not only selection failures. Even when only the annotated evidence sentences are supplied under their two source titles, the reader must align entities and relations across scripts, preserve question language, and compose the two facts. The result motivates dedicated reader adaptation rather than assuming that multilingual instruction tuning already solves mixed-language reasoning. The 8B screening row (21.62/32.12) and the independently recorded matched-transfer run below (21.94/32.23) differ slightly; the surviving provenance does not identify a prompt, parser, or decoding cause. We preserve both historical values, name the runs separately, and compute deltas only within the matched comparison table.

\subsection{RQ1: direct transfer versus translate-train adaptation}

Table~\ref{tab:module-results} provides the historically reported matched module comparisons. Direct one-shot Llama~3.1~8B reaches 32.23 reader F1. English-task tuning does not transfer automatically: the English Bactrainus reader reaches 28.48 F1, 3.75 points below direct prompting. In contrast, \xreader\ reaches 53.41 F1, a gain of 21.18 points (65.7\% relative) over direct prompting and 24.93 points (87.5\% relative) over English transfer. EM follows the same direction, rising from 21.94 to 39.85. The complete \xreader\ and English-selector prediction files are not present in the local bundle, so these differences are historical evidence rather than a new artifact-level recomputation.

\begin{table}[htbp]
\caption{Historically reported matched transfer aggregates on $n=7{,}405$ under legacy Hotpot normalization. Reader evaluation uses annotated supporting sentences; selector evaluation uses all candidates.}
\label{tab:module-results}
\centering
\small
\begin{tabular}{@{}llrr@{}}
\toprule
Task & System & EM & F1 \\
\midrule
\multirow{4}{*}{Reader}
& Llama 3.1 8B, matched one-shot & 21.94 & 32.23 \\
& English Bactrainus & 14.39 & 28.48 \\
& \xreader & 39.85 & 53.41 \\
& \xreader\ + 70B rationale & \textbf{41.73} & \textbf{55.62} \\
\midrule
\multirow{2}{*}{Selector}
& English Bactrainus & 20.27 & 43.69 \\
& \xselector & \textbf{57.42} & \textbf{84.23} \\
\bottomrule
\end{tabular}
\end{table}

The negative English-transfer difference should not be read as evidence that task tuning is intrinsically harmful. It is consistent with a distribution-narrowing explanation: the English module learned a useful selection/answering policy whose lexical access layer remained specialized to English, whereas the direct instruction model retained broader multilingual behavior. Translate-train adaptation exposes both the task and language interfaces jointly and recovers the lost behavior. The surviving complete selector output independently re-scores to 57.29 support EM and 84.42 F1 under corrected title--sentence keys, close to the historical 57.42/84.23 row.

\subsection{RQ2: selection shows the larger observed adaptation gain}

Within each task, the selector difference is larger than the reader difference. \xselector\ improves support F1 from 43.69 to 84.23, a 40.54-point gain (92.8\% relative), and support EM from 20.27 to 57.42. Reader F1 improves by 24.93 points over its English-transfer counterpart. This asymmetry is compatible with the language assignment probabilities in Eqs.~\ref{eq:mix-prob}--\ref{eq:distinct-languages}: the selector must search an average of more than eight candidate languages, while the reader sees only the annotated supporting sentences. Because answer F1 and support F1 measure different outputs, the gain sizes do not by themselves prove that selection is causally harder; they identify where adaptation is associated with the larger within-task change.

Formally, the end-to-end probability of a correct answer decomposes as
\begin{align}
\Pr(Y=1)&=\Pr(E=1)\Pr(Y=1\mid E=1)\\
&\quad+\Pr(E=0)\Pr(Y=1\mid E=0),
\label{eq:error-decomp}
\end{align}
where $E$ denotes recovery of the complete gold evidence set. When $\Pr(Y=1\mid E=0)$ is small for connected questions, improvements in $\Pr(E=1)$ have a multiplicative effect on pipeline reliability. Supporting-fact EM rising to 57.42 still leaves 42.58\% of items without an exact evidence set, which contributes to joint performance remaining far below oracle-reader F1.

\subsection{End-to-end composition}

Table~\ref{tab:e2e} confirms the module analysis. With the English selector, changing only the reader from English Bactrainus to one-shot Llama~8B or Llama~70B raises answer F1, but joint F1 remains below 20 because support F1 is fixed at 43.69. Replacing both modules with translate-train variants increases answer F1 to 50.97 and joint F1 to 40.12. The rationale continuation adds 0.65 answer-F1 and 0.37 joint-F1 points, reaching 51.62 and 40.49.

\begin{table}[htbp]
\caption{Historically reported end-to-end selector--reader aggregates on $n=7{,}405$. Complete aligned end-to-end predictions are not present in the audited local bundle. Support values are identical for rows sharing a selector.}
\label{tab:e2e}
\centering
\scriptsize
\setlength{\tabcolsep}{3pt}
\begin{tabular}{@{}llrrrrrr@{}}
\toprule
& & \multicolumn{2}{c}{Support} & \multicolumn{2}{c}{Answer} & \multicolumn{2}{c}{Joint} \\
\cmidrule(lr){3-4}\cmidrule(lr){5-6}\cmidrule(l){7-8}
Selector & Reader & EM & F1 & EM & F1 & EM & F1 \\
\midrule
English Bact. & English Bact. & 20.27 & 43.69 & 10.17 & 18.48 & 5.54 & 12.91 \\
English Bact. & Llama 8B & 20.27 & 43.69 & 15.32 & 21.32 & 8.93 & 15.60 \\
English Bact. & Llama 70B & 20.27 & 43.69 & 20.11 & 29.92 & 11.12 & 19.92 \\
\xselector & \xreader & 57.42 & 84.23 & 36.87 & 50.97 & 21.64 & 40.12 \\
\xselector & \xreader\ + rat. & \textbf{57.42} & \textbf{84.23} & \textbf{37.04} & \textbf{51.62} & \textbf{22.03} & \textbf{40.49} \\
\bottomrule
\end{tabular}
\end{table}

The best end-to-end answer F1 is only 4.00 points below the corresponding oracle-evidence reader F1 (55.62). This comparatively small answer gap coexists with an 11.13-point gap between answer F1 and joint F1. The aggregate gap is compatible with answer credit being earned without exact sentence support, but only aligned per-example outputs can quantify such cases. They may reflect alternative valid evidence, annotation granularity, parametric knowledge, or unsupported guessing; per-example inspection is needed to distinguish them.

\begin{figure}[htbp]
\centering
\includegraphics[width=0.98\linewidth]{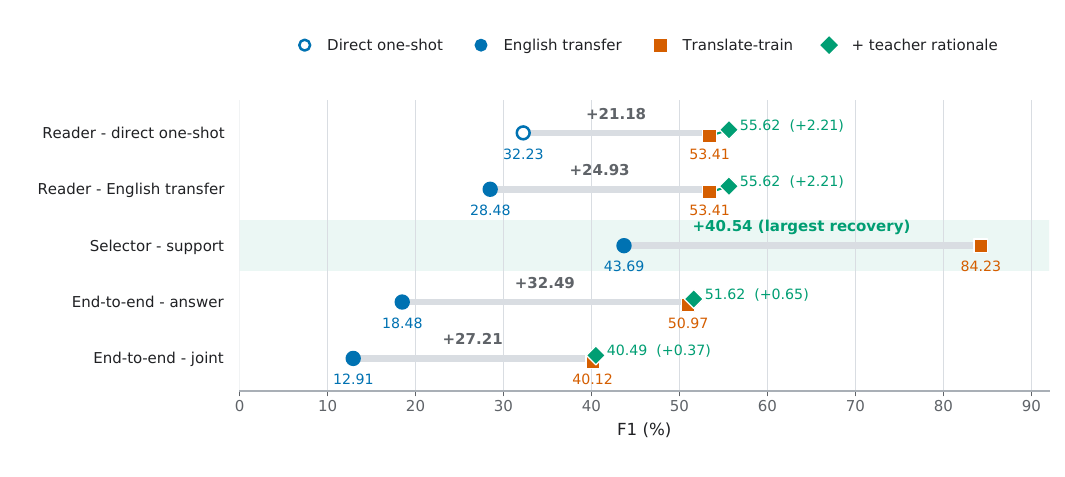}
\caption{Historical F1 changes from English/direct transfer to translate-train adaptation. The largest reported within-task change occurs in supporting-fact selection, while teacher rationales add a smaller increment after cross-lingual adaptation. Different metric components are not directly commensurate.}
\label{fig:transfer}
\end{figure}
\FloatBarrier

\subsection{RQ3: limited incremental effect of teacher rationales}

The rationale continuation improves oracle-reader F1 from 53.41 to 55.62 (2.21 points) but end-to-end answer F1 by only 0.65 points. A plausible explanation is that English rationales help the reader organize already selected facts but do not change the selector, language placement, or translation. Another is a ceiling imposed by translation and answer normalization. Because the continuation changes training examples and duration simultaneously, the result does not identify the causal effect of chain-of-thought supervision. A controlled study should hold token budget, epochs, and examples constant and compare rationales in the question language, evidence languages, and English.

\subsection{RQ4: realized question--paragraph language effects}

We aligned every validation question, answer, reconstructed reader context, and selector gold set to its recovered HotpotQA ID. Complete oracle-evidence results survive for the historically labeled Llama~3.1~70B reader, GPT-4o mini, Qwen2~72B, and the adapted selector; other systems in Tables~\ref{tab:screen}--\ref{tab:e2e} are not used in this item-level analysis. We recompute answer F1 with the Unicode/script-aware protocol and support F1 with paragraph-title plus sentence-index keys. For disjoint conditions $A$ and $B$, the descriptive contrast is
\begin{equation}
\widehat\Delta_M(A,B)=\frac{1}{|A|}\sum_{i\in A}M(x_i)
-\frac{1}{|B|}\sum_{i\in B}M(x_i),
\label{eq:condition-contrast}
\end{equation}
with percentile intervals from nonparametric item bootstraps (fixed analysis seed 20260810). The primary S2/S4 and script contrasts in Table~\ref{tab:language-effects} use 10,000 replicates; the exploratory role and distractor contrasts use 2,000. These are associations under one realized random assignment, not model-independent causal effects.

Table~\ref{tab:language-effects} shows a stable reader pattern. Relative to S2, where exactly one of two gold paragraphs matches $\ell_q$, S4 places both gold paragraphs outside $\ell_q$ and in different languages. S4 answer F1 is lower by 15.79 points for Llama, 10.25 for GPT-4o mini, and 14.05 for Qwen; none of the reader intervals includes zero. In contrast, the selector changes by only $-1.71$ points and its interval includes zero. Different-script rather than same-script gold evidence produces still larger reader gaps (11.98--23.70 points), whereas the selector gap is $-1.78$. The assignment geometry is therefore associated with oracle-evidence answer composition much more sharply than with candidate selection in these frozen outputs.

\begin{table}[htbp]
\caption{Artifact-verified F1 by realized language condition. S2/S3/S4 contain 564/312/6,515 items. Contrasts are percentage points with 95\% item-bootstrap intervals; reader rows use answer F1 and the selector row uses support F1. ``Llama 3.1 70B'' is the historical artifact label, not an independently verified endpoint identity.}
\label{tab:language-effects}
\centering
\scriptsize
\setlength{\tabcolsep}{3pt}
\begin{tabular}{@{}lrrrll@{}}
\toprule
System & S2 & S3 & S4 & $\Delta$(S4$-$S2) & $\Delta$(different$-$same script) \\
\midrule
Llama 3.1 70B & 60.60 & 39.60 & 44.81 & $-15.79$ [$-19.41,-12.12$] & $-19.79$ [$-22.58,-16.99$] \\
GPT-4o mini & 57.79 & 47.34 & 47.54 & $-10.25$ [$-13.75,-6.83$] & $-11.98$ [$-14.70,-9.25$] \\
Qwen2 72B & 53.83 & 36.58 & 39.78 & $-14.05$ [$-17.63,-10.46$] & $-23.70$ [$-26.45,-20.91$] \\
\xselector & 85.93 & 86.26 & 84.22 & $-1.71$ [$-3.60,0.21$] & $-1.78$ [$-3.20,-0.32$] \\
\bottomrule
\end{tabular}
\end{table}

\begin{figure}[htbp]
\centering
\includegraphics[width=0.98\linewidth]{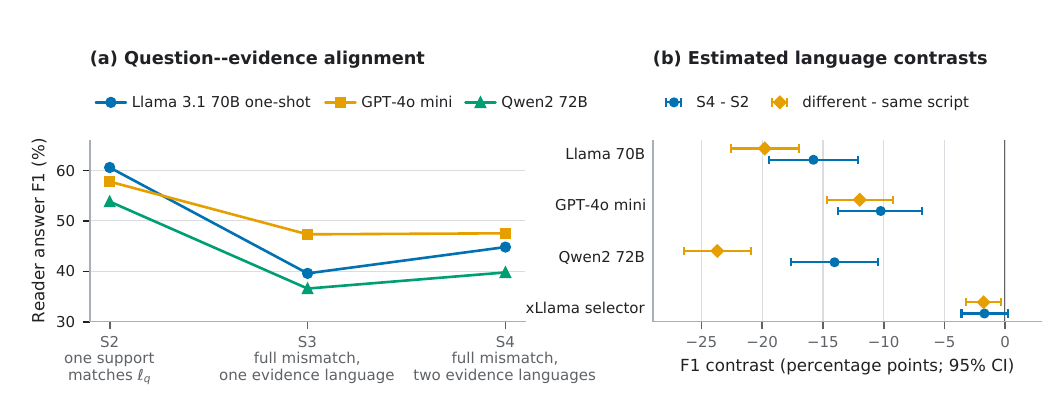}
\caption{Language-conditioned performance from 7,405-row prediction artifacts. (a) Reader answer F1 across the populated S2--S4 strata. (b) S4-minus-S2 and different-minus-same-script contrasts with 95\% item-bootstrap intervals. Values are descriptive; group composition is not held fixed.}
\label{fig:language-effects}
\end{figure}
\FloatBarrier

Role labels sharpen the interpretation for 5,010 bridge items in which the annotated source-language support sentences under one gold title uniquely contain the normalized answer. Matching the assigned answer-hop paragraph language to $\ell_q$ is associated with $+19.69$ [$13.61,25.15$], $+14.43$ [$8.98,20.04$], and $+21.51$ [$15.75,27.03$] reader-F1 points for Llama, GPT-4o mini, and Qwen, respectively. A narrower comparison of bridge-hop mismatch ($\ell_q=\ell_t\ne\ell_b$, $n=193$) against answer-hop mismatch ($\ell_q=\ell_b\ne\ell_t$, $n=197$) favors the former by 6.06 [$-2.33,14.48$], 3.80 [$-3.88,11.34$], and 15.01 [$6.85,23.06$] points; only Qwen's interval excludes zero. Answer-hop alignment is therefore a strong correlate of reader F1, but the present base split does not support a universal ranking of bridge and answer-hop difficulty. Supplementary Section~B.4 reports denominators and all exploratory contrasts.

The assigned gold-paragraph languages add a different view. Using $G_{\ell}(x)$ from Eq.~\ref{eq:gold-language-membership}, marginal F1 ranges from 34.6 (Thai) to 51.1 (French) for Llama, 43.7 (Urdu) to 52.4 (French) for GPT-4o mini, 36.4 (Turkish) to 45.4 (Hindi) for Qwen, and 81.8 (Korean) to 86.1 (Greek) for the selector. Because these overlapping margins confound question language, script, role, and the other assigned gold language, we also fit a one-row-per-source OLS sensitivity model that jointly represents both gold languages and controls question-language fixed effects, question type, answer type, script relation, exact question--gold alignment, same-language pairing, support-fact count, and structural flags. Relative to substituting one English-assigned gold paragraph, only Llama retains FDR-significant answer-F1 associations: Thai $-13.42$ [$-17.96,-8.87$], Korean $-12.92$ [$-17.60,-8.24$], Mandarin Chinese $-11.89$ [$-16.54,-7.24$], and Urdu $-6.42$ [$-11.03,-1.81$]. No gold-language coefficient survives within-outcome Benjamini--Hochberg correction for GPT-4o mini, Qwen, or the selector. The evidence therefore supports model-specific associations, not a universal hierarchy of gold-evidence languages. Supplementary Section~B.3 gives the complete margins, intervals, composition counts, and role caveats.

Distractor analyses tell a different story because only the selector observes non-gold paragraphs. Having at least one distractor in $\ell_q$ changes selector F1 by $+0.51$ points (95\% CI $[-0.63,1.58]$); having a distractor in a support language changes it by $-0.76$ [$-1.72,0.23$]; and $K_C\ge9$ rather than $K_C\le8$ changes it by $+0.56$ [$-0.49,1.58$]. These intervals include zero. The earlier 40.54-point historical translate-train gain therefore describes adaptation relative to an English selector, whereas the realized within-model contrasts show little residual sensitivity to how distractor languages are arranged. Adaptation gain and conditional language sensitivity are distinct quantities.

The complete \datasetplus\ design would permit the paired estimand
\begin{equation}
\Delta_M(\ell,\ell')=\frac{1}{N}\sum_{i=1}^{N}
\left[M(x_i^{\ell})-M(x_i^{\ell'})\right],
\label{eq:paired-language}
\end{equation}
where evidence is fixed and only question--answer language changes. Until the missing parallel validation views are generated, Eq.~\ref{eq:paired-language} remains a prespecified V2 analysis rather than a reported result. Supplementary Section~E provides the corresponding covariate and multiplicity controls.

\subsection{RQ5: V2 RC1 improves paragraph and question judgments, not answers}\label{sec:judge-results}

\begin{figure}[htbp]
\centering
\includegraphics[width=0.98\linewidth]{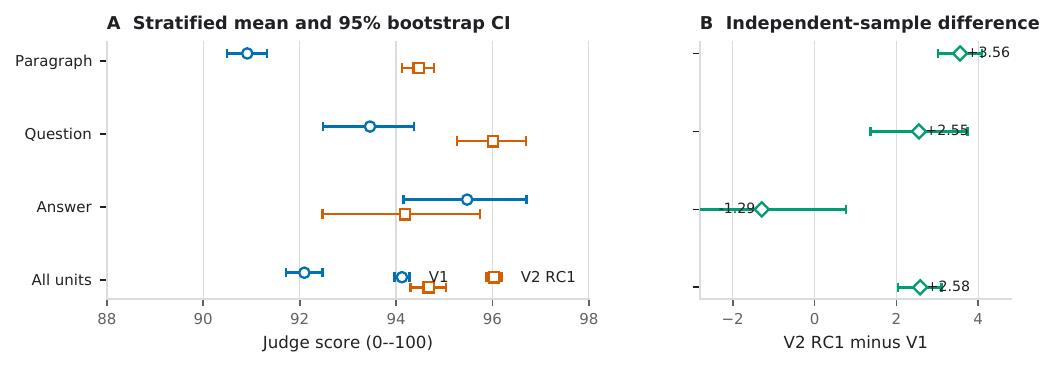}
\caption{Balanced source-referenced translation audit. Panel A gives independently bootstrapped version means; Panel B gives V2 RC1-minus-V1 differences from independent language-by-unit-stratified resamples. The comparison is descriptive and unpaired.}
\label{fig:judge-summary}
\end{figure}

Table~\ref{tab:judge-summary} and Figure~\ref{fig:judge-summary} report the independent source-referenced audit. The pooled V2 RC1 mean is 94.68 versus 92.10 for V1, a descriptive difference of $+2.58$ points (95\% independent stratified-bootstrap CI $[2.04,3.10]$). Paragraphs account for most sampled units and have a $+3.56$ [$3.02,4.09$] difference; questions have $+2.55$ [$1.37,3.73$]. These intervals exclude zero under the fixed sampling design, but they remain conditional on one prompt, one requested judge alias, and independent rather than matched content.

\begin{table}[htbp]
\caption{Source-referenced GLM-5.2 judge scores. Each version contains 1,840 paragraphs, 460 questions, and 460 answers, balanced within 23 target languages. Brackets are 95\% independent language-by-unit-stratified bootstrap intervals (20,000 replicates); differences are V2 RC1 minus V1.}
\label{tab:judge-summary}
\centering
\small
\begin{tabular}{@{}lrrr@{}}
\toprule
Unit & V1 mean [CI] & V2 RC1 mean [CI] & Difference [CI] \\
\midrule
Paragraph & 90.91 [90.49, 91.32] & 94.47 [94.13, 94.79] & $+3.56$ [3.02, 4.09] \\
Question & 93.46 [92.49, 94.37] & 96.01 [95.27, 96.69] & $+2.55$ [1.37, 3.73] \\
Answer & 95.47 [94.15, 96.71] & 94.18 [92.47, 95.75] & $-1.29$ [$-3.38$, 0.77] \\
All units & 92.10 [91.71, 92.47] & 94.68 [94.30, 95.03] & $+2.58$ [2.04, 3.10] \\
\bottomrule
\end{tabular}
\end{table}

\FloatBarrier

Answers do not follow the aggregate pattern: their mean difference is $-1.29$ [$-3.38,0.77$], and V2 RC1 has greater answer-score dispersion (sample SD 19.46 versus 14.47). Exact English answer copies occur in 179/460 V2 audit answers versus 116/460 V1 answers. They are not uniformly erroneous: names, formulas, acronyms, and numbers may conventionally remain unchanged. Indeed, non-copy V2 answers average 97.21, while exact-copy V2 answers average 89.44; the corresponding V1 means are 94.27 and 99.05. Lower V2 answer scores are concentrated in script-sensitive cases rather than showing a uniform loss of factual adequacy. Thai answer scores fall by 30.70 points, Mandarin Chinese by 13.75, and Greek by 8.15, even though Thai and Chinese paragraph/question scores rise. This heterogeneity, detailed in Supplementary Section~D, is why we do not summarize RC1 as simply ``higher quality.''

Coverage and defect attribution are separate from these sampled judge scores. The candidate covers 99.00\% of the intended sources. Malformed HotpotQA records triggered both validation omissions, while the V2 workflow's decision to omit rather than preserve and flag them remains a downstream treatment; Supplementary Section~C separates these cases from clean-source training-generation omissions.

\section{Discussion}\label{sec:discussion}

\subsection{Implications for knowledge-based reasoning systems}

The central design decision is to attach language to evidence units rather than to an example as a whole. In a conventional translated benchmark, the language variable is one categorical label. In \dataset, it is the vector in Eq.~\ref{eq:language-vector}. This makes two systems with the same aggregate F1 distinguishable: one may fail whenever $\ell_b\ne\ell_t$, while another may be robust to cross-hop composition but vulnerable to distractors written in the question language. Such differences are scientifically meaningful and actionable for retrieval, prompting, or training.

The design also clarifies what \dataset\ does not measure. It conditions on a small supplied candidate set, so it removes multilingual indexing, corpus coverage, and first-stage retrieval recall. This is a feature for diagnosis, because the selector's input is controlled, but a limitation for deployment. A complete multilingual RAG evaluation should pair \dataset-like evidence composition with an open-corpus retrieval resource such as XOR QA or XRAG.

\subsection{Evidence fidelity matters more than answer-only gains}

The best pipeline reaches 51.62 answer F1 but 40.49 joint F1. An answer-only interpretation would miss the remaining evidence gap. Sentence support serves at least three purposes: it reveals selector failure, supports error attribution, and discourages unsupported answers from being counted as complete successes. At the same time, exact support labels are not logically exhaustive; a system may find an alternative correct chain. A future release should permit adjudicated alternative evidence or evaluate entailment from predicted evidence while retaining the original labels for comparability.

The selector's larger historically reported adaptation gain suggests a practical priority. In a modular pipeline, improving a reader under oracle evidence does not repair errors introduced upstream by candidate selection. At the same time, Table~\ref{tab:language-effects} finds much larger language-arrangement gaps for readers than for the adapted selector. These are compatible results: the first compares training regimes, whereas the second compares inputs within one frozen system. Together they motivate evaluating candidate scoring, entity normalization, and hop-conditioned reading alongside the final generator without treating cross-task F1 differences as causal estimates of difficulty.

\subsection{Cross-lingual composition is the principal observed sensitivity}

The strongest diagnostic result is not simply that multilingual inputs are difficult, but that the location of the language boundary matters. Moving from partial question--gold alignment to complete mismatch is associated with a 10.25--15.79 point reader-F1 deficit across the three verified reader artifacts, compared with 1.71 points for the adapted selector. Likewise, different-script gold evidence is associated with reader deficits of 11.98--23.70 points but only 1.78 points for the selector. Under the supplied-candidate design, candidate selection is therefore comparatively stable while answer generation remains sensitive to aligning entities and composing propositions across language and script boundaries.

This contrast is descriptive rather than a causal comparison between modules: the selector and readers have different outputs, losses, and training histories. Moreover, S0 has no observed items and S1 has only 14, so the reported contrasts measure severity and location of crossing within a highly mixed benchmark, not a cross-lingual-versus-monolingual penalty. Nevertheless, the pattern recurs under two independently defined language geometries and survives exclusion of structurally flagged validation items (Supplementary Section~C). This convergence motivates a concrete systems hypothesis: future gains may require bridge-entity normalization and hop-aware cross-lingual composition, not only stronger relevance scoring.

\subsection{Scale, multilinguality, and reasoning are not interchangeable}

The one-shot screen shows a broad association between model capability and reader F1 but no monotonic rule across families. A large model can have strong general reasoning but uneven lexical coverage; a smaller multilingual model can retrieve cross-lingual correspondences yet fail at connected reasoning; an instruction-tuned endpoint can obey answer-language constraints better than a larger open checkpoint. Three component tests are therefore needed: single-hop access in each language, cross-language bridge transfer, and composition with both hops. Aggregate model size is not a substitute.

\subsection{Translation quality is role- and language-dependent}

The V2 audit illustrates why a single corpus-level translation score is insufficient. Paragraph and question means move upward across most languages, while answer behavior is dominated by a smaller number of names, titles, acronyms, and script conventions. The sharp Thai, Chinese, and Greek answer regressions co-occur with more exact English copies, yet exact copying is not itself a semantic error: retaining \emph{FIFA}, a chemical formula, or a personal name in Latin script can be conventional, whereas leaving an ordinary phrase untranslated is not. The judge prompt explicitly exempts conventionally Latin-script names, but its explanations apply that exception unevenly across languages. The per-language table and score distributions should therefore be read as a triage map for bilingual review, not a ranking of languages or a conclusive V1--V2 preference.

\subsection{Cross-script and typological analysis}

The inventory covers ten writing-system groups, so script must be modeled separately from language family. Persian and Urdu share an Arabic-derived script but differ from Arabic linguistically; Hindi and Urdu are closely related but use different dominant scripts; Japanese combines several writing systems; and Chinese, Japanese, and Thai require non-whitespace metrics. In the three 7,405-row reader artifacts, different-script gold evidence is associated with 11.98--23.70 fewer F1 points than same-script evidence; the adapted selector gap is 1.78 points. Persian has the lowest reader F1 in all three verified systems while its selector F1 remains near the macro average. These observations make the covered scripts informative diagnostics, but they do not by themselves distinguish orthography, training exposure, translation quality, morphology, or entity transliteration.

\subsection{Versioned regeneration}

Resource maintenance should not overwrite earlier translations. Each translator therefore defines a new version over stable source IDs and a recorded language assignment. V2 is one Gemma~4~31B instantiation of this rule, not a retroactive correction of V1. Comparing versions requires the same stratified bilingual audit and explicit completeness checks; model recency alone is not evidence of uniformly better translation.

\section{Limitations and threats to validity}\label{sec:limitations}

\textbf{Translation-derived content.} The benchmark inherits English Wikipedia's entity and cultural distribution. Fluent translations do not create native information needs, and translation quality may vary by language, script, entity type, and sentence length.

\textbf{Random language mixture.} Independent assignment gives strong coverage and closed-form expectations, but it does not guarantee equal counts for every ordered question--bridge--answer language triple. Role-balanced paired subsets are needed for fine-grained comparisons.

\textbf{Legacy assignment and provenance.} V1 did not record the assignment seed for its historical base-train draw. The public V1.1 release freezes a deterministic one-view projection and labels it as a packaging choice, not as reconstruction of the lost draw. V2 uses a separately realized assignment and therefore cannot support a paired translator comparison with V1.

\textbf{Structural defects and incomplete resources.} Neither historical V1 nor the V2 candidate is error-free, and the parallel validation half of \datasetplus\ is unavailable. In V2, inherited HotpotQA defects trigger both validation omissions, while most missing training instances are clean-source omissions from an incomplete generation run. Excluding upstream-malformed rows is nevertheless a workflow treatment, not an inevitable consequence of the source anomaly. Supplementary Section~C reports the full attribution and sensitivity analysis; the main results should not be interpreted as certifying either artifact as semantically correct in every language.

\textbf{No realized monolingual control.} The intended assignment produces no S0 observations and only 14 S1 items in validation. Condition contrasts therefore compare degrees and locations of cross-lingual mixing; they do not estimate the effect of introducing cross-linguality relative to an otherwise matched monolingual control.

\textbf{Model-judge validity.} The source-referenced translation audit uses one requested model alias, one rubric, and one score per unit, without bilingual human calibration or repeated judge models. The stored configuration does not authenticate the provider-resolved GLM checkpoint. Independent samples prevent paired content control, and bootstrap intervals omit judge and endpoint uncertainty. Script and transliteration conventions can also be applied inconsistently, especially for short named-entity answers.

\textbf{Reasoning necessity.} HotpotQA support annotations do not prove that each item requires both hops for every model. Parametric knowledge, shortcuts, or one sufficient paragraph can produce a correct answer. Eq.~\ref{eq:necessity} and partial-evidence evaluation should accompany claims about reasoning.

\textbf{Fixed candidates.} Selector scores do not include corpus-scale retrieval. Results cannot establish multilingual indexing quality, retrieval latency, or robustness to missing gold documents.

\textbf{Metrics across scripts.} Exact match and token F1 can penalize valid aliases, morphology, segmentation, and transliteration. Character F1 and semantic review help but introduce their own biases. Per-language human analysis remains necessary.

\textbf{Development-set estimation.} There is no hidden multilingual test set, and the 7,405 validation instances support model/recipe comparison. Reported maxima may therefore include selection on the development set. A future release should freeze a held-out test subset or use a public evaluation server.

\textbf{Single-run baselines.} Training results are point estimates. Language-pair analysis multiplies comparisons and requires paired uncertainty, several seeds, and correction such as Holm's procedure.

\textbf{Artifact coverage.} Complete item-level files survive for three one-shot readers and one selector; absent responses are scored as empty. Transfer and end-to-end tables remain historical aggregate results until their complete prediction files are deposited.

\textbf{Evidence roles.} Bridge and answer-hop order is well defined for many bridge questions but not all comparison questions. Role-aware analyses must preserve an ``unordered comparison'' category rather than force a bridge/target interpretation.

\section{Conclusion}\label{sec:conclusion}

\dataset\ turns cross-lingual multi-hop QA into a resource with explicit linguistic structure. Its transformation assigns language independently to the question--answer pair and candidate paragraphs while retaining the fixed-candidate reasoning roles and source support keys of HotpotQA. The formalization separates question/evidence interface crossing from multilingual evidence composition and supplies measurable mismatch, entropy, richness, and script descriptors. The audit also shows why transformation fidelity must be measured rather than assumed.

Across the 7,405-row prediction artifacts, full question--evidence mismatch is associated with 10.25--15.79 lower reader F1 than partial alignment, and different-script evidence with 11.98--23.70 lower F1 than same-script evidence. Marginal scores also vary with the assigned gold-paragraph language, but after joint covariate adjustment and within-outcome FDR correction, significant language coefficients remain for only one reader; neither the other two readers nor the selector supports a universal gold-language hierarchy. The corresponding adapted-selector alignment gaps are small, and direct distractor-language contrasts have intervals spanning zero. Historical all-split experiments additionally report translate-train reader/support F1 of 53.41/84.23 and a strongest end-to-end answer/joint F1 of 51.62/40.49; these remain historical aggregates pending deposit of their full predictions.

The source-referenced audit adds a secondary result: the Gemma-generated V2 candidate receives higher paragraph and question judgments but no supported answer improvement; lower Thai, Chinese, and Greek answer scores co-occur with more script-sensitive exact copies. HotpotQA defects trigger its two validation omissions, while clean-source omissions from the incomplete training run still prevent V2 from replacing V1. The benchmark therefore complements, rather than supersedes, same-language translations, controlled small probes, and open-retrieval multilingual RAG datasets. Its value lies in broad language coverage, within-instance role variables, distractors, explainable supervision, and explicit version boundaries. A completed V2 and complete \datasetplus\ release require full coverage and bilingual review rather than model recency alone.

\section*{CRediT authorship contribution statement}
\textbf{Iman Barati:} Conceptualization, Methodology, Data curation, Software, Investigation, Formal analysis, Writing -- original draft. \textbf{Arash Ghafouri:} Supervision. \textbf{Behrouz Minaei-Bidgoli:} Supervision, Validation. All authors reviewed and approved the final manuscript and are accountable for the work.

\section*{Funding}
This research did not receive any specific grant from funding agencies in the public, commercial, or not-for-profit sectors.

\section*{Declaration of competing interest}
The authors declare that they have no known competing financial interests or personal relationships that could have appeared to influence the work reported in this paper.

\section*{Data availability}
The \href{https://github.com/Iman998/XhotpotQA}{XHotpotQA code repository}, \href{https://huggingface.co/datasets/Iman998/XhotpotQA}{audited V1.1 dataset}, and \href{https://huggingface.co/datasets/Iman998/XhotpotQA-V2}{Gemma-generated V2 candidate} are publicly available. The source-referenced \href{https://huggingface.co/datasets/Iman998/XhotpotQA-GLM52-Judge-V1}{V1 judge sample} and \href{https://huggingface.co/datasets/Iman998/XhotpotQA-GLM52-Judge-V2}{V2 judge sample} are cross-linked from the dataset cards as one release family. Dataset files inherit CC BY-SA 4.0 from HotpotQA; repository software uses MIT.

\section*{Ethics statement}
This study transforms a publicly available benchmark and involves no intervention, private records, or collection of personal or sensitive data. Three specialists provided task-level judgments of machine-translated text during translator selection; no participant characteristics were collected, and the judgments were analyzed only as expert assessment of research artifacts. Formal research ethics approval and informed consent for publication of personal data were therefore not applicable.

\section*{Declaration of generative AI and AI-assisted technologies in the manuscript preparation process}
During preparation of this work, the authors used OpenAI Codex for drafting support, language editing, and LaTeX restructuring. After using this tool, the authors reviewed and edited the content as needed and take full responsibility for the content of the publication. Models used as experimental translators, rationale generators, readers, selectors, or judges are research methods and are reported separately in the corresponding methodology sections.

\bigskip
\section*{Supplementary Material}
\addcontentsline{toc}{section}{Supplementary Material}
The following appendices report the full language distributions, metric and gold-language sensitivity analyses, resource-quality checks, source-referenced judge diagnostics, statistical specifications, exact prompts, and evaluated model identifiers.
\appendix
\section{Full language distribution}\label{app:distribution}

Table~\ref{tab:language-distribution} reports marginal assignment counts. ``Support'' denotes supporting-fact annotation occurrences rather than unique paragraphs; ``Docs'' denotes candidate paragraph occurrences. Because paragraphs are assigned independently, question counts alone do not characterize question--evidence pairs.

\begingroup
\small
\setlength{\tabcolsep}{5pt}
\begin{longtable}{@{}lrrrrrr@{}}
\caption{Language distribution in the historical base-train projection and audited legacy validation assignment.}\label{tab:language-distribution}\\
\toprule
& \multicolumn{3}{c}{Train} & \multicolumn{3}{c}{Validation} \\
\cmidrule(lr){2-4}\cmidrule(l){5-7}
Language & Questions & Support & Docs & Questions & Support & Docs \\
\midrule
\endfirsthead
\toprule
& \multicolumn{3}{c}{Train} & \multicolumn{3}{c}{Validation} \\
\cmidrule(lr){2-4}\cmidrule(l){5-7}
Language & Questions & Support & Docs & Questions & Support & Docs \\
\midrule
\endhead
English & 677 & 1,579 & 6,591 & 309 & 804 & 3,107 \\
Mandarin Chinese & 644 & 1,579 & 6,472 & 294 & 716 & 3,126 \\
Hindi & 661 & 1,535 & 6,606 & 315 & 727 & 3,144 \\
Spanish & 684 & 1,623 & 6,647 & 289 & 780 & 3,048 \\
Arabic & 608 & 1,609 & 6,482 & 339 & 742 & 3,125 \\
French & 609 & 1,497 & 6,349 & 276 & 718 & 3,022 \\
Bengali & 717 & 1,583 & 6,501 & 292 & 673 & 3,111 \\
Portuguese & 670 & 1,545 & 6,523 & 314 & 721 & 2,948 \\
Russian & 679 & 1,578 & 6,530 & 333 & 735 & 2,994 \\
Urdu & 629 & 1,591 & 6,494 & 324 & 804 & 3,139 \\
Indonesian & 665 & 1,616 & 6,547 & 310 & 732 & 2,962 \\
German & 597 & 1,586 & 6,498 & 336 & 719 & 3,028 \\
Japanese & 624 & 1,599 & 6,453 & 305 & 768 & 3,129 \\
Turkish & 633 & 1,575 & 6,411 & 310 & 736 & 3,038 \\
Vietnamese & 634 & 1,497 & 6,431 & 274 & 761 & 3,020 \\
Swahili & 633 & 1,605 & 6,562 & 304 & 746 & 3,001 \\
Korean & 633 & 1,530 & 6,471 & 315 & 807 & 3,054 \\
Persian & 664 & 1,640 & 6,469 & 295 & 819 & 3,105 \\
Italian & 724 & 1,587 & 6,404 & 299 & 736 & 3,109 \\
Thai & 683 & 1,604 & 6,519 & 320 & 831 & 3,154 \\
Dutch & 642 & 1,488 & 6,541 & 314 & 736 & 3,204 \\
Polish & 654 & 1,528 & 6,484 & 316 & 723 & 3,099 \\
Greek & 616 & 1,610 & 6,470 & 316 & 754 & 2,992 \\
Swedish & 681 & 1,641 & 6,533 & 306 & 717 & 3,041 \\
\midrule
\textbf{Total} & \textbf{15,661} & \textbf{37,825} & \textbf{155,988}
& \textbf{7,405} & \textbf{18,005} & \textbf{73,700} \\
\bottomrule
\end{longtable}
\endgroup

\section{Artifact-verified metric and language sensitivity}\label{app:metric-sensitivity}

Throughout this appendix, ``Llama 70B'' and ``Llama 3.1 70B'' denote the historical reader-artifact label. The surviving client does not independently establish the checkpoint served by the private endpoint; resolving that identity requires the server log.

\subsection{Normalization protocol}

Table~\ref{tab:metric-sensitivity} re-scores the three 7,405-row reader artifacts under both the historical Hotpot normalizer and the release protocol. Unicode normalization and character tokenization for Chinese, Japanese, and Thai increase aggregate F1 by 2.78--3.21 points while leaving EM nearly unchanged. The largest per-language F1 changes occur for Chinese (22.26--26.05 points), Japanese (24.11--28.73), and Thai (13.70--17.20). Consequently, model comparisons within one protocol are meaningful, but cross-paper language rankings are not comparable unless tokenization is named.

\begin{table}[htbp]
\caption{Metric sensitivity on 7,405-row reader artifacts. Missing responses are scored as empty; values are percentages.}
\label{tab:metric-sensitivity}
\centering
\small
\begin{tabular}{@{}lrrrrr@{}}
\toprule
& \multicolumn{2}{c}{Legacy Hotpot} & \multicolumn{2}{c}{Unicode/script aware} & $\Delta$F1 \\
\cmidrule(lr){2-3}\cmidrule(lr){4-5}
System & EM & F1 & EM & F1 &  \\
\midrule
Llama 3.1 70B & 30.84 & 42.81 & 30.68 & 45.83 & +3.02 \\
GPT-4o mini & 29.94 & 45.15 & 29.71 & 48.35 & +3.21 \\
Qwen2 72B & 24.94 & 37.95 & 24.79 & 40.73 & +2.78 \\
\bottomrule
\end{tabular}
\end{table}

\subsection{Question-language results}

Table~\ref{tab:per-language-results} reports release-normalized F1 for the same readers and corrected support F1 for the complete selector artifact. Reader dispersion is substantial: the observed ranges are 29.99--65.56 (Llama), 34.18--57.81 (GPT-4o mini), and 13.19--62.32 (Qwen), whereas selector F1 occupies the much narrower 81.21--87.53 interval. Persian is the minimum reader language for all three systems (29.99/34.18/13.19) while its selector F1 is 84.25. This convergence is descriptive rather than a claim about Persian itself: translation adequacy, entity transliteration, training exposure, script, and answer morphology are not experimentally separated.

\begingroup
\footnotesize
\renewcommand{\arraystretch}{0.86}
\setlength{\tabcolsep}{4pt}
\begin{longtable}{@{}lrrrrr@{}}
\caption{Artifact-verified performance by question language. Reader columns are Unicode/script-aware answer F1; selector is support F1.}\label{tab:per-language-results}\\
\toprule
Language & $n$ & Llama 70B & GPT-4o mini & Qwen2 72B & \xselector \\
\midrule
\endfirsthead
\toprule
Language & $n$ & Llama 70B & GPT-4o mini & Qwen2 72B & \xselector \\
\midrule
\endhead
Arabic & 339 & 33.52 & 42.37 & 23.29 & 85.32 \\
Bengali & 292 & 34.34 & 36.41 & 21.32 & 84.87 \\
German & 336 & 51.19 & 55.83 & 56.86 & 82.41 \\
Greek & 316 & 37.51 & 39.21 & 26.72 & 85.51 \\
English & 309 & 65.56 & 50.97 & 62.32 & 87.53 \\
Spanish & 289 & 53.06 & 56.34 & 54.56 & 81.21 \\
Persian & 295 & 29.99 & 34.18 & 13.19 & 84.25 \\
French & 276 & 57.18 & 56.69 & 57.14 & 84.30 \\
Hindi & 315 & 42.50 & 45.81 & 27.10 & 85.12 \\
Indonesian & 310 & 51.16 & 54.26 & 49.74 & 85.53 \\
Italian & 299 & 50.00 & 52.93 & 56.60 & 84.06 \\
Japanese & 305 & 57.47 & 55.40 & 40.26 & 83.59 \\
Korean & 315 & 32.49 & 36.74 & 18.53 & 83.48 \\
Dutch & 314 & 50.07 & 53.93 & 55.57 & 86.13 \\
Polish & 316 & 50.12 & 54.47 & 51.91 & 85.86 \\
Portuguese & 314 & 55.44 & 57.81 & 53.77 & 86.28 \\
Russian & 333 & 40.07 & 45.77 & 33.92 & 85.39 \\
Swedish & 306 & 48.62 & 50.12 & 55.96 & 83.91 \\
Swahili & 304 & 41.53 & 40.41 & 32.56 & 84.41 \\
Thai & 320 & 44.91 & 50.60 & 35.22 & 81.69 \\
Turkish & 310 & 48.35 & 52.33 & 50.87 & 84.43 \\
Urdu & 324 & 34.50 & 41.27 & 22.33 & 84.02 \\
Vietnamese & 274 & 46.25 & 50.67 & 46.04 & 83.86 \\
Mandarin Chinese & 294 & 46.53 & 46.80 & 35.07 & 82.60 \\
\midrule
Macro-language & -- & 45.93 & 48.39 & 40.87 & 84.41 \\
\bottomrule
\end{longtable}
\endgroup

\subsection{Gold-paragraph-language results}\label{app:gold-language}

Each validation item contributes one unordered pair of assigned gold-paragraph languages: 326 pairs (4.40\%) use one language twice, 2,006 (27.09\%) use two languages written in the same script, and 5,073 (68.51\%) cross scripts. Table~\ref{tab:gold-language-marginal} reports the overlapping strata $G_{\ell}(x)=1$. Their counts are not additive, and their differences are not pairwise interventions; each interval resamples source items rather than paragraph rows.

\begin{table}[htbp]
\caption{Performance when a language is assigned to at least one gold paragraph. Reader columns report Unicode-aware answer F1; the selector column reports supporting-fact F1. Brackets are 95\% source-item bootstrap intervals (2,000 replicates). An item with two different assigned gold languages enters both relevant marginal rows, but never more than once within a row; consequently, the language-row counts are not additive.}
\label{tab:gold-language-marginal}
\centering
\scriptsize
\setlength{\tabcolsep}{3pt}
\resizebox{\linewidth}{!}{%
\begin{tabular}{llrcccc}
\toprule
Language & Script & $n$ & Llama-3-70B & GPT-4o-mini & Qwen2-72B & Selector \\
\midrule
English & Latin & 642 & 49.5 [46.1, 52.9] & 51.0 [47.9, 54.2] & 43.2 [39.9, 46.5] & 83.5 [81.6, 85.3] \\
Mandarin Chinese & Han & 583 & 36.4 [33.1, 40.2] & 45.5 [42.1, 49.1] & 38.8 [35.4, 42.0] & 83.7 [81.7, 85.5] \\
Spanish & Latin & 632 & 49.2 [45.8, 52.5] & 48.5 [45.2, 51.7] & 38.6 [35.4, 41.8] & 85.6 [84.0, 87.2] \\
Hindi & Devanagari & 596 & 44.7 [41.2, 48.4] & 50.1 [46.8, 53.4] & 45.4 [42.0, 49.0] & 83.5 [81.6, 85.4] \\
Arabic & Arabic & 602 & 43.3 [40.0, 46.8] & 46.2 [42.9, 49.6] & 41.6 [38.5, 44.9] & 84.9 [83.1, 86.6] \\
French & Latin & 583 & 51.1 [47.6, 54.7] & 52.4 [48.8, 56.1] & 39.7 [36.3, 43.1] & 84.1 [82.1, 86.1] \\
Russian & Cyrillic & 588 & 46.1 [42.6, 49.6] & 45.0 [41.7, 48.5] & 39.1 [36.0, 42.3] & 84.3 [82.4, 86.1] \\
Portuguese & Latin & 571 & 50.6 [46.9, 54.0] & 50.3 [46.9, 53.6] & 41.5 [38.0, 44.9] & 83.5 [81.6, 85.4] \\
Bengali & Bengali & 548 & 50.7 [47.2, 54.2] & 48.3 [44.7, 51.8] & 43.9 [40.5, 47.2] & 84.0 [82.0, 85.9] \\
Urdu & Arabic & 627 & 42.4 [39.2, 45.9] & 43.7 [40.6, 47.0] & 42.4 [39.2, 45.5] & 84.6 [82.7, 86.3] \\
Indonesian & Latin & 585 & 47.6 [44.0, 51.2] & 50.4 [46.9, 53.9] & 39.2 [35.7, 42.4] & 86.0 [84.2, 87.7] \\
Japanese & Japanese & 613 & 45.5 [42.0, 49.0] & 49.0 [45.8, 52.3] & 40.2 [37.0, 43.4] & 84.1 [82.4, 85.8] \\
German & Latin & 592 & 47.3 [43.8, 50.8] & 48.5 [45.0, 51.9] & 37.7 [34.3, 41.1] & 85.5 [83.7, 87.3] \\
Swahili & Latin & 599 & 48.1 [44.7, 51.7] & 48.3 [45.0, 51.7] & 39.8 [36.5, 43.3] & 84.5 [82.7, 86.3] \\
Turkish & Latin & 591 & 44.7 [41.3, 48.4] & 46.2 [42.9, 49.5] & 36.4 [33.3, 39.9] & 84.0 [82.2, 85.7] \\
Vietnamese & Latin & 615 & 48.3 [44.8, 51.7] & 48.3 [45.2, 51.5] & 42.8 [39.5, 46.1] & 85.3 [83.7, 86.9] \\
Korean & Hangul & 630 & 35.2 [31.8, 38.8] & 44.7 [41.4, 47.8] & 38.7 [35.8, 41.9] & 81.8 [79.9, 83.5] \\
Italian & Latin & 601 & 48.1 [44.5, 51.7] & 49.4 [46.2, 52.8] & 40.7 [37.2, 44.1] & 86.0 [84.3, 87.6] \\
Persian & Arabic & 639 & 43.7 [40.2, 47.1] & 45.9 [42.6, 49.1] & 44.7 [41.4, 47.8] & 82.8 [81.1, 84.5] \\
Thai & Thai & 663 & 34.6 [31.3, 38.0] & 47.8 [44.8, 51.0] & 40.6 [37.6, 43.8] & 82.9 [81.1, 84.6] \\
Dutch & Latin & 593 & 50.2 [46.7, 53.5] & 48.9 [45.7, 52.4] & 40.3 [36.9, 43.6] & 84.1 [82.3, 85.8] \\
Polish & Latin & 599 & 50.2 [46.6, 53.8] & 51.5 [48.1, 54.7] & 40.3 [36.9, 43.7] & 85.3 [83.4, 87.0] \\
Greek & Greek & 611 & 48.3 [44.9, 51.6] & 51.8 [48.4, 55.1] & 42.6 [39.2, 45.9] & 86.1 [84.4, 87.7] \\
Swedish & Latin & 581 & 49.1 [45.6, 52.7] & 49.4 [46.3, 52.8] & 41.0 [37.7, 44.4] & 85.4 [83.6, 87.2] \\
\bottomrule
\end{tabular}%
}
\end{table}

For the 5,010 bridge items with mechanically determinable roles, the archived analysis further reports each language separately in bridge and answer-hop position with source-item bootstrap intervals. The direction of answer-versus-bridge differences is not stable across models or languages, so we do not infer a general answer-hop-language penalty. The adjusted sensitivity model described in the main text uses both gold-document language counts jointly, English as the omitted reference, and HC3 standard errors. Benjamini--Hochberg correction is applied separately within each system--metric outcome. Its coefficients are conditional descriptive substitutions, not causal effects: translation quality remains unobserved, and the legacy assignment seed was not recorded. Machine-readable marginal, role, unordered-pair, script-pair, adjusted-coefficient, diagnostic, and input-provenance files accompany the source package.

\subsection{Role and distractor contrasts}\label{app:role-effects}

Table~\ref{tab:role-distractor-effects} collects the exploratory contrasts discussed in RQ4. Reader contrasts use Unicode/script-aware answer F1; selector contrasts use corrected support F1. The role comparison is restricted to bridge questions for which the annotated source support sentences under one gold title uniquely contain the normalized answer. The positive and negative group sizes are reported because these conditions are neither balanced nor paired. All intervals in this table use 2,000 stratified item-bootstrap replicates.

\begin{table}[htbp]
\caption{Exploratory role- and distractor-language contrasts in percentage-point F1.}
\label{tab:role-distractor-effects}
\centering
\scriptsize
\setlength{\tabcolsep}{3pt}
\begin{tabularx}{\linewidth}{@{}lYrrr@{}}
\toprule
System & Positive group minus negative group & $n_+$ & $n_-$ & $\Delta$ [95\% CI] \\
\midrule
Llama 3.1 70B & answer-hop language aligned vs. not & 205 & 4,805 & $+19.69$ [$13.61,25.15$] \\
GPT-4o mini & answer-hop language aligned vs. not & 205 & 4,805 & $+14.43$ [$8.98,20.04$] \\
Qwen2 72B & answer-hop language aligned vs. not & 205 & 4,805 & $+21.51$ [$15.75,27.03$] \\
Llama 3.1 70B & bridge-hop mismatch vs. answer-hop mismatch & 193 & 197 & $+6.06$ [$-2.33,14.48$] \\
GPT-4o mini & bridge-hop mismatch vs. answer-hop mismatch & 193 & 197 & $+3.80$ [$-3.88,11.34$] \\
Qwen2 72B & bridge-hop mismatch vs. answer-hop mismatch & 193 & 197 & $+15.01$ [$6.85,23.06$] \\
\xselector & any question-language distractor vs. none & 2,134 & 5,271 & $+0.51$ [$-0.63,1.58$] \\
\xselector & distractor shares a gold language vs. none & 3,509 & 3,896 & $-0.76$ [$-1.72,0.23$] \\
\xselector & $K_C\ge9$ vs. $K_C\le8$ & 3,314 & 4,091 & $+0.56$ [$-0.49,1.58$] \\
\bottomrule
\end{tabularx}
\end{table}

\section{Resource-quality and release audit}\label{app:quality-details}

\subsection{Validity dimensions and automatic gates}

An incorrect translation can mimic a reasoning failure. If a bridge entity is transliterated inconsistently, the model faces an entity-linking problem that the source instance did not contain. If polarity changes in a yes/no question, the gold answer becomes wrong. If sentence boundaries change, the supporting index points to a different proposition. We distinguish five validity dimensions:

\begin{enumerate}[leftmargin=*]
\item \emph{semantic adequacy}: the translated unit preserves source meaning;
\item \emph{answer preservation}: question and answer retain the original relation and polarity;
\item \emph{entity/number preservation}: named entities, dates, quantities, and comparison direction remain consistent;
\item \emph{structural preservation}: paragraph and sentence cardinality/order remain unchanged; and
\item \emph{language compliance}: output is in the assigned language and uses an acceptable script/variant.
\end{enumerate}

The release toolkit verifies non-empty output, Unicode validity, sentence cardinality, identifier stability, supporting-fact bounds, content integrity, and question/answer language equality. These gates describe the corrected format; they were not all enforced by the historical generator. Language identification, duplicate-content review, named-entity/number consistency, and answer occurrence belong to the versioned audit protocol; they are not inferred from structural validation alone. Automatic back-translation is useful for triage but cannot establish semantic equivalence.

\subsection{V1 structural audit and release decision}

We joined all 7,405 translated validation rows to the original HotpotQA source order and compared every translated paragraph against its source sentence array. Table~\ref{tab:structural-audit} reports the audit encoded in the public V1.1 status fields; V1.1 also carries the corresponding original English \texttt{source\_sentences}. In total, 424 of 73,700 paragraph occurrences change sentence cardinality (418 shortfalls and six surpluses), affecting 407 instances. Thirty-three instances contain a blank translated sentence, and nine gold support indices fall outside the shortened translated array. The historical reader-construction notebook silently omitted those nine support facts, so the defect changes the evaluated input rather than merely its metadata. The union of canonical-release-blocking structural flags is 439 instances (5.93\%). Those rows are retained with \texttt{quarantined} status in audited V1.1; four additional items contain duplicate normalized translated titles and are retained as \texttt{review\_required}.

\begin{table}[htbp]
\caption{Structural audit of the raw V1 validation archive. Occurrences count paragraph/fact events; affected items are deduplicated within each row.}
\label{tab:structural-audit}
\small
\begin{tabularx}{\linewidth}{@{}YrrY@{}}
\toprule
Finding & Occurrences & Items & Canonical action \\
\midrule
Sentence-cardinality mismatch & 424 & 407 & regenerate or adjudicate paragraph \\
Blank translated sentence & 33 & 33 & regenerate sentence/paragraph \\
Support index out of range & 9 & 9 & block label until alignment is repaired \\
Union of blocking flags & -- & 439 & retain as quarantined; correct for canonical V2 \\
\bottomrule
\end{tabularx}
\end{table}

This audit does not establish semantic adequacy for the remaining items. It does, however, test whether the flagged subset drives the reported aggregate behavior. Excluding all 439 flagged instances changes release-normalized F1 by $+0.10$ (Llama), $-0.12$ (GPT-4o mini), $+0.11$ (Qwen), and $+0.24$ (selector); the largest absolute change over the tested EM/F1 metrics is 0.51 points. Table~\ref{tab:quality-condition-sensitivity} additionally recomputes the two principal language-condition contrasts after the same exclusion. The largest absolute contrast shift is 0.22 points, and every paired shift interval includes zero. Aggregate and condition-level conclusions are therefore numerically insensitive to this exclusion, but a corrected V2 still requires repaired text and a versioned correction record rather than silent deletion or overwrite.

\begin{table}[htbp]
\caption{Sensitivity of primary F1 contrasts to excluding the 439 structurally flagged validation items. ``All'' uses 7,405 sources and ``clean'' uses 6,966; shift is clean minus all with a paired 95\% source-item-bootstrap interval (2,000 replicates). The Llama name is the historical artifact label.}
\label{tab:quality-condition-sensitivity}
\centering
\scriptsize
\setlength{\tabcolsep}{3pt}
\begin{tabular}{@{}lllll@{}}
\toprule
System & Contrast & All & Clean & Shift [95\% CI] \\
\midrule
Llama 3.1 70B & S4$-$S2 & $-15.79$ & $-15.82$ & $-0.03$ [$-0.94,0.85$] \\
GPT-4o mini & S4$-$S2 & $-10.25$ & $-10.03$ & $+0.22$ [$-0.65,1.08$] \\
Qwen2 72B & S4$-$S2 & $-14.05$ & $-13.92$ & $+0.13$ [$-0.75,0.98$] \\
\xselector & S4$-$S2 & $-1.71$ & $-1.54$ & $+0.17$ [$-0.18,0.50$] \\
\addlinespace
Llama 3.1 70B & Different$-$same script & $-19.79$ & $-19.90$ & $-0.11$ [$-0.81,0.62$] \\
GPT-4o mini & Different$-$same script & $-11.98$ & $-12.18$ & $-0.20$ [$-0.94,0.53$] \\
Qwen2 72B & Different$-$same script & $-23.70$ & $-23.57$ & $+0.13$ [$-0.60,0.79$] \\
\xselector & Different$-$same script & $-1.78$ & $-1.76$ & $+0.02$ [$-0.35,0.39$] \\
\bottomrule
\end{tabular}
\end{table}

\subsection{V2 RC1 completeness and source preservation}\label{app:v2-audit}

We reconciled the supplied V2 files against the official HotpotQA hard-training and distractor-development source orders instead of inferring completeness from successful parsing. The candidate contains 22,836 of 23,066 intended instances (99.00\%). Every retained instance has a unique source ID and stores English source-sentence arrays that exactly reproduce the corresponding HotpotQA arrays alongside the translated candidates. The 230 absent instances, however, prevent a complete row-level audit and rule out a canonical-release claim.

The official source itself is not structurally perfect. It contains 137 records with a blank-string sentence element and five whose annotated supporting index lies outside the source sentence array. There are no zero-length source paragraph arrays: the relevant emptiness is a blank sentence inside an otherwise populated paragraph. Calling every downstream anomaly an XHotpotQA translation failure would therefore be incorrect.

Table~\ref{tab:v2-defect-attribution} classifies the missing and retained anomalies by the earliest verifiable origin. Of the 230 missing rows, 50 correspond to source records with a blank sentence or out-of-bounds support annotation: 48 in training (44 blank-string records and all four out-of-bounds records) and both missing validation rows. The first missing validation source contains an empty final sentence in a distractor paragraph; the second contains supporting-fact index 902 for a five-sentence paragraph. Thus, upstream HotpotQA defects trigger both validation omissions; the fact that the V2 workflow omitted rather than preserved and flagged those rows is still a downstream treatment. The remaining 180 missing rows all belong to the training split and have clean source structure; they are generation/orchestration omissions introduced in the V2 workflow. Among retained records, the other 92 source blank strings are represented by 92 em-dash placeholders (44 train, 48 validation). This transformation is attributable to XHotpotQA even though the triggering emptiness is inherited; the source text remains recoverable in the stored source array. A separate exact-string screen finds 1,612 training and 837 validation non-English sentence outputs identical to the English source. Those 2,449 matches are \emph{review signals}, not automatically defects: punctuation-only strings, numbers, formulas, and proper names may legitimately be invariant, while an untranslated ordinary sentence may not.

\begin{table}[htbp]
\caption{V2 RC1 anomaly attribution. ``Source-origin'' identifies the earliest triggering defect in HotpotQA; row exclusion remains the V2 workflow's treatment. Both validation omissions are source-triggered, whereas clean-source omissions occur only in training.}
\label{tab:v2-defect-attribution}
\centering
\small
\begin{tabularx}{\linewidth}{@{}YrrrY@{}}
\toprule
Finding & Train & Validation & Total & Attribution and release treatment \\
\midrule
Official-source blank-string sentence elements & 88 & 49 & 137 & inherited source anomaly; preserve and flag \\
Official-source support index out of bounds & 4 & 1 & 5 & inherited label anomaly; adjudicate upstream \\
Missing RC1 rows with either source anomaly & 48 & 2 & 50 & inherited trigger; both validation omissions fall here \\
Missing RC1 rows with clean source structure & 180 & 0 & 180 & XHotpotQA generation/orchestration omission \\
Retained source blanks changed to em dash & 44 & 48 & 92 & XHotpotQA transformation of inherited blank \\
Non-English sentence exactly equals English source & 1,612 & 837 & 2,449 & triage signal; requires content-aware review \\
\bottomrule
\end{tabularx}
\end{table}

The retry ledger alone does not establish completeness. A late retry process stopped before recording an outcome for every queued instance, after which filtered retries operated only on the smaller recorded-error set. The final ledger therefore named three unresolved records, whereas source-order reconciliation finds 230 absent IDs. Source-ID reconciliation is consequently the authoritative count. This run-history limitation says nothing about the semantic validity of retained translations, and it must not obscure that both validation omissions originate upstream in HotpotQA.

\noindent\textbf{Release decision.} V2 is suitable for public analysis as an \emph{audited incomplete snapshot}. It improves auditability by retaining exact English source sentences, but it is not yet a complete replacement for V1. Completion requires restoring the 230 absent source IDs, representing inherited blank strings without silent semantic substitution, adjudicating the five inherited support-label anomalies, and repeating the alignment and bilingual quality review.

\subsection{Future paired acceptance target}

A future translator-effect study should use the frozen V1 target-language assignment, not merely the same source IDs. The supplied RC1 does not meet that condition, so Eqs.~\ref{eq:v2-structural-quality}--\ref{eq:v2-human-quality} are an acceptance target rather than an analysis of the current judge samples. Let $j=(i,u,\ell)$ denote a matched translation unit with source ID $i$, stable unit ID $u$, and target language $\ell$, and let $\mathcal J_i$ be the set of units belonging to source $i$. Define $e_i^{(v)}=1$ when any $j\in\mathcal J_i$ has a release-blocking structural defect in version $v\in\{1,2\}$. The instance-level structural defect rate and paired change are
\begin{equation}
R_{\mathrm{struct}}^{(v)}=\frac{1}{N}\sum_{i=1}^{N}e_i^{(v)},
\qquad
\Delta R_{2-1}=R_{\mathrm{struct}}^{(2)}-R_{\mathrm{struct}}^{(1)}.
\label{eq:v2-structural-quality}
\end{equation}
A negative $\Delta R_{2-1}$ is necessary but not sufficient: a structurally valid mistranslation can still change meaning. For human-audit score $h_{jr}^{(v)}\in[1,5]$ on matched unit $j$ from reviewer $r$, define the source-clustered paired improvement
\begin{equation}
\Delta H_{2-1}=\frac{1}{|\mathcal A|}\sum_{i\in\mathcal A}
\frac{1}{|\mathcal J_i|}\sum_{j\in\mathcal J_i}
\left(\frac{1}{|\mathcal R_j|}\sum_{r\in\mathcal R_j}
h_{jr}^{(2)}-h_{jr}^{(1)}\right),
\label{eq:v2-human-quality}
\end{equation}
where $\mathcal A$ is the stratified bilingual source sample. Source-clustered bootstrap intervals accompany $\Delta H_{2-1}$; paired binary preservation flags use matched-pair intervals or McNemar tests. A future claim that canonical V2 has higher quality requires complete key equality across versions, zero release-blocking structural defects in the corrected files, positive bilingual adequacy evidence under Eq.~\ref{eq:v2-human-quality}, and no material regression in answer/entity/language preservation. The current independent GLM audit does not meet these paired-identification requirements, and model recency alone is not an acceptance criterion.

\subsection{Human audit design}

The translator-selection pilot provides comparative evidence for one choice, not corpus-level quality estimates. A release-grade bilingual audit stratifies by all 24 languages, question type, answer type, support role, script, and mismatch condition. Each sampled unit receives 1--5 adequacy and fluency scores plus binary labels for answer, entity/number, sentence-alignment, and assigned-language preservation. Two bilingual annotators review each item; disagreements are adjudicated by a third reviewer. Agreement is reported per criterion, and cluster-bootstrap confidence intervals account for several judgments from one source instance.

The primary audit unit is the complete instance, not an isolated sentence, because independently fluent sentences can still translate a bridge entity inconsistently. Errors receive one of four consequences: \emph{cosmetic}, \emph{metric-sensitive}, \emph{reasoning-changing}, or \emph{label-invalidating}. Only the first category may remain uncorrected in the corrected canonical V2; audited V1.1 instead retains all categories with explicit status and flags for reproducibility. Corrected text retains the original generation record and a versioned adjudication trace.

\subsection{Data statement and intended use}

The source questions were authored for English HotpotQA and refer predominantly to English Wikipedia content. Target-language text is machine translated rather than elicited from speakers. The resource therefore measures model behavior under controlled linguistic transformations; it does not measure what speakers in each community naturally ask, which knowledge sources they prefer, or how code-switching occurs in practice. The language inventory is broad in scripts and families but not balanced by speaker population, region, gender, dialect, or socioeconomic context. No new personal information is collected beyond public Wikipedia-derived content and the original HotpotQA annotations.

Intended uses are benchmark research, controlled ablation, multilingual evidence selection, model transfer, and evaluation-tool development. Out-of-scope uses include ranking languages or communities, estimating human reasoning ability, deploying high-stakes QA without independent validation, or treating translation-derived scores as a measure of cultural coverage.

\section{Source-referenced translation-judge audit details}\label{app:judge-details}

\subsection{Audit-set construction}

Table~\ref{tab:judge-artifacts} makes the analysis population explicit. The V1 file is already the completed balanced sample. The initial V2 file contains the complete paragraph and question samples but only 240 answer rows, all created with the English question in \texttt{source\_text}; they do not judge answer translation. We discard all 240, including the one row without a parseable score, and replace the answer stratum with the separate 460-row corrected run. No statistic in this article uses the invalid answer rows. The final V2 analysis population is thus the logical concatenation of 1,840 paragraphs, 460 questions, and 460 corrected answers. Both versions have 80/20/20 units in every target language.

\begin{table}[htbp]
\caption{Construction of the two balanced judge samples. P, Q, and A denote paragraph, question, and answer units.}
\label{tab:judge-artifacts}
\centering
\small
\begin{tabularx}{\linewidth}{@{}YrrrY@{}}
\toprule
Stage & Paragraphs & Questions & Answers & Use in analysis \\
\midrule
V1 balanced sample & 1,840 & 460 & 460 & All 2,760 units retained \\
V2 preliminary run & 1,840 & 460 & 240 & P/Q retained; all preliminary A rows discarded \\
V2 corrected-answer run & -- & -- & 460 & Replaces the preliminary A stratum \\
V2 balanced sample & 1,840 & 460 & 460 & All 2,760 final units retained \\
\bottomrule
\end{tabularx}
\end{table}

The requested endpoint model string is only \texttt{glm-5.2}. The public Z.ai model card and GLM-5 technical report identify an intended public family \citep{glm5team2026glm5,zai2026glm52}, but neither proves which provider checkpoint served this private alias. We therefore report ``requested GLM-5.2 alias,'' not an immutable model revision. The evaluator produces one source-referenced score per unit; comments in the historical module that mention two reference/reference-free scores do not describe the serialized records and are not used here.

\subsection{Aggregate distributions}

Table~\ref{tab:judge-distributions} complements the means with dispersion and quantiles. V2 paragraph and question distributions shift upward and have fewer scores below 80. Answer medians and interquartile ranges remain at 100 because short factual answers often receive a ceiling score, while a small low-scoring tail increases the V2 answer SD. This ceiling-heavy distribution is another reason not to treat the $-1.29$ mean difference as a uniform answer degradation.

\begin{table}[htbp]
\caption{Judge-score distributions. SD is the sample standard deviation; $<80$ is the percentage below 80.}
\label{tab:judge-distributions}
\centering
\small
\begin{tabular}{@{}llrrrrrrr@{}}
\toprule
Version & Unit & $n$ & Mean & SD & Q1 & Median & Q3 & $<80$ \\
\midrule
V1 & Paragraph & 1,840 & 90.91 & 9.47 & 89 & 94 & 96 & 9.89\% \\
V1 & Question & 460 & 93.46 & 10.71 & 92 & 96 & 100 & 7.61\% \\
V1 & Answer & 460 & 95.47 & 14.47 & 100 & 100 & 100 & 6.09\% \\
V1 & All & 2,760 & 92.10 & 10.81 & 90 & 95 & 100 & 8.88\% \\
\addlinespace
V2 RC1 & Paragraph & 1,840 & 94.47 & 7.53 & 94 & 96 & 98 & 3.48\% \\
V2 RC1 & Question & 460 & 96.01 & 8.12 & 95 & 98 & 100 & 4.13\% \\
V2 RC1 & Answer & 460 & 94.18 & 19.46 & 100 & 100 & 100 & 6.74\% \\
V2 RC1 & All & 2,760 & 94.68 & 10.59 & 95 & 97 & 100 & 4.13\% \\
\bottomrule
\end{tabular}
\end{table}

\begin{figure}[htbp]
\centering
\includegraphics[width=0.98\linewidth]{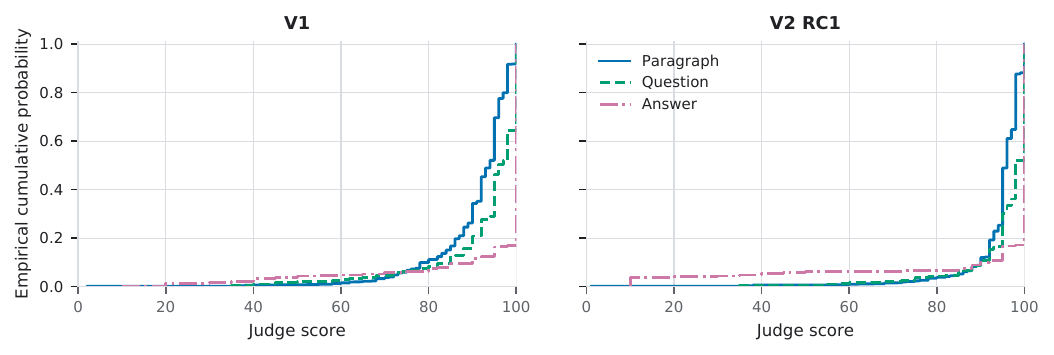}
\caption{Empirical score distributions for the independently sampled V1 and V2 RC1 audit units. The discrete mass at 100 is especially large for short answers; ECDFs therefore reveal tail behavior hidden by medians.}
\label{fig:judge-ecdf}
\end{figure}
\FloatBarrier

\subsection{Per-language results and targeted interpretation}

\begin{figure}[p]
\centering
\includegraphics[width=0.86\linewidth,height=0.78\textheight,keepaspectratio]{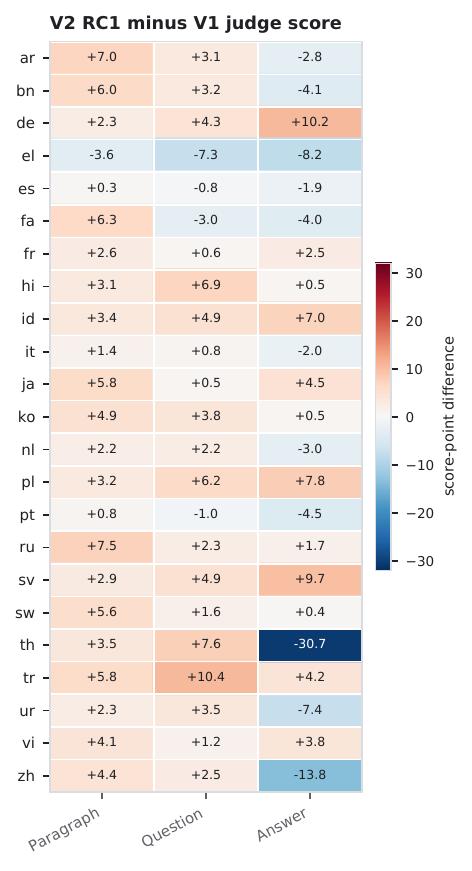}
\caption{Per-language V2 RC1-minus-V1 mean differences under independent balanced samples. Cells are descriptive; they are not paired translator effects or language rankings.}
\label{fig:judge-language-heatmap}
\end{figure}

Table~\ref{tab:judge-by-language} reports every target language. Because each row has the same 80:20:20 unit allocation, its overall mean is comparable across rows within this audit; it is not a population-weighted estimate of real-world language use. The heat map in Figure~\ref{fig:judge-language-heatmap} emphasizes that the pooled gain is not homogeneous.

\begin{longtable}{@{}lrrrrrr@{}}
\caption{Per-language source-referenced judge means and V2 RC1-minus-V1 differences. P, Q, and A denote paragraph, question, and answer. Overall means use the fixed 80:20:20 weighting.}\label{tab:judge-by-language}\\
\toprule
Lang. & $\Delta$P & $\Delta$Q & $\Delta$A & V1 overall & V2 overall & $\Delta$overall \\
\midrule
\endfirsthead
\multicolumn{7}{l}{\tablename\ \thetable\ (continued)}\\
\toprule
Lang. & $\Delta$P & $\Delta$Q & $\Delta$A & V1 overall & V2 overall & $\Delta$overall \\
\midrule
\endhead
\midrule
\multicolumn{7}{r}{Continued on next page}\\
\endfoot
\bottomrule
\endlastfoot
ar & +6.98 & +3.10 & $-2.85$ & 91.03 & 95.72 & +4.69 \\
bn & +6.04 & +3.20 & $-4.10$ & 91.32 & 95.19 & +3.88 \\
de & +2.29 & +4.35 & +10.25 & 92.79 & 96.75 & +3.96 \\
el & $-3.58$ & $-7.30$ & $-8.15$ & 90.81 & 85.85 & $-4.96$ \\
es & +0.35 & $-0.75$ & $-1.90$ & 95.39 & 95.18 & $-0.21$ \\
fa & +6.30 & $-2.95$ & $-4.00$ & 90.93 & 93.98 & +3.04 \\
fr & +2.56 & +0.55 & +2.50 & 94.94 & 97.16 & +2.22 \\
hi & +3.11 & +6.90 & +0.50 & 92.03 & 95.34 & +3.31 \\
id & +3.39 & +4.85 & +7.00 & 92.76 & 96.99 & +4.23 \\
it & +1.36 & +0.75 & $-2.05$ & 95.31 & 96.00 & +0.69 \\
ja & +5.75 & +0.50 & +4.50 & 91.33 & 96.00 & +4.67 \\
ko & +4.90 & +3.75 & +0.50 & 92.98 & 96.95 & +3.98 \\
nl & +2.21 & +2.25 & $-3.00$ & 92.90 & 94.25 & +1.35 \\
pl & +3.23 & +6.20 & +7.75 & 91.53 & 96.00 & +4.47 \\
pt & +0.81 & $-1.00$ & $-4.50$ & 95.87 & 95.49 & $-0.38$ \\
ru & +7.49 & +2.30 & +1.65 & 89.64 & 95.29 & +5.65 \\
sv & +2.89 & +4.90 & +9.65 & 89.90 & 94.25 & +4.35 \\
sw & +5.56 & +1.65 & +0.40 & 89.38 & 93.43 & +4.05 \\
th & +3.54 & +7.65 & $-30.70$ & 91.41 & 89.93 & $-1.48$ \\
tr & +5.84 & +10.40 & +4.15 & 89.18 & 95.50 & +6.32 \\
ur & +2.29 & +3.55 & $-7.40$ & 92.32 & 93.20 & +0.88 \\
vi & +4.07 & +1.25 & +3.85 & 92.60 & 96.17 & +3.57 \\
zh & +4.40 & +2.55 & $-13.75$ & 91.89 & 92.96 & +1.07 \\
\end{longtable}

\clearpage

\FloatBarrier

\clearpage

Three patterns warrant review rather than post-hoc generalization. First, Greek is the only language with a large negative overall difference ($-4.96$), and all three unit means decrease. Low-scoring RC1 Greek paragraphs include mixed English intrusions and corrupted lexical material, so this row cannot be explained only by short-answer conventions. Second, Thai paragraphs and questions increase by 3.54 and 7.65 points, but answers decrease by 30.70. Twelve of 20 audited RC1 Thai answers exactly copy the English answer, compared with none in the V1 Thai sample; those 12 receive a mean of 48.0. Third, Chinese paragraphs and questions increase by 4.40 and 2.55 while answers decrease by 13.75. Three RC1 Chinese answers exactly copy English and average 43.33; the corresponding V1 sample has one exact copy scored 100. Greek RC1 has ten exact-copy answers (mean 76.5) versus two in V1 (both scored 100). These small denominators and different sampled items preclude a causal model comparison, but they identify high-priority bilingual audit strata.

Across all answer languages, exact source copies rise from 116/460 in V1 to 179/460 in V2 RC1. Their means fall from 99.05 to 89.44, whereas non-copy answers rise from 94.27 to 97.21. The split is diagnostic, not a validity rule. The answer rubric says conventionally Latin-script names should not be penalized and minor transliteration differences are acceptable, yet the retained explanations sometimes score an unchanged personal name as a wrong-language copy in one script community and as conventional in another. Conversely, an unchanged multiword common-noun answer can genuinely violate the target-language requirement. Dataset release should therefore combine exact-copy and script detection with entity typing and bilingual adjudication; it should never auto-reject every invariant string.

\section{Stratified analysis specification}\label{app:analysis}

\subsection{Reporting matrix}

For a metric $M$, the macro-language average used in the main article is
\begin{equation}
\operatorname{MacroLang}(M)=\frac{1}{|\mathcal L|}
\sum_{\ell\in\mathcal L}\frac{1}{N_{\ell}}
\sum_{x:\ell_q=\ell}M(x).
\label{eq:supp-macro}
\end{equation}
The main aggregate should be accompanied by the following disaggregation for answer, support, and joint EM/F1:

\begin{enumerate}[leftmargin=*]
\item question language and macro-language average (Eq.~\ref{eq:supp-macro});
\item ordered question--gold language pair and balanced pair macro-average;
\item populated S0--S4 mismatch conditions, $\rho_G$, $\rho_D$ (with no-distractor items marked NA), $H_G$, $H_C$, and $K_C$;
\item bridge/comparison type and bridge/answer-hop role where defined;
\item same-script, mixed-script, and different-script evidence;
\item same-family versus different-family language pairs;
\item entity, numeric, date, and yes/no answer types;
\item context length, number of support facts, and number of distinct candidate languages; and
\item human-audit status and translation error category.
\end{enumerate}

The balanced pair macro-average for metric $M$ is
\begin{equation}
\operatorname{MacroPair}(M)=\frac{1}{|\mathcal P|}
\sum_{(\ell_q,\ell_g)\in\mathcal P}
\frac{1}{N_{\ell_q,\ell_g}}
\sum_{x\in\mathcal X_{\ell_q,\ell_g}}M(x),
\label{eq:macro-pair}
\end{equation}
where $\ell_g$ may denote the answer-hop language or an unordered evidence-language set. This prevents high-count pairs from masking rare directions.

\subsection{Regression and uncertainty}

For binary EM in the base validation split, where each source appears once, a prespecified logistic model can test conditional associations:
\begin{align}
\operatorname{logit}\Pr(Y_i=1)=&\ \beta_0+\beta_1\rho_G(i)+\beta_2H_G(i)+\beta_3\rho_D(i)\\
&+\beta_4\rho_G(i)H_G(i)+\gamma_{\ell_q(i)}+\delta_{t(i)}
+\eta_{a(i)}.
\label{eq:mixed-model}
\end{align}
Here $t(i)$ is question type and $a(i)$ answer type. The 21 no-distractor items are omitted from terms involving $\rho_D$ or modeled with an explicit missingness indicator. Translation-audit quality and length enter as sensitivity covariates. A source random intercept is added only for repeated \datasetplus\ views; it is unidentified and unnecessary when the base split has one observation per source. For bounded F1, we prefer item bootstrap estimates---clustered by source for repeated views---to an unqualified Gaussian model.

Pairwise system comparisons resample source IDs and report the distribution of
\begin{equation}
\widehat\Delta_{A,B}=\frac{1}{N}\sum_i\left[M_A(x_i)-M_B(x_i)\right].
\label{eq:paired-bootstrap}
\end{equation}
In a confirmatory evaluation, Holm correction should control family-wise error across prespecified language/condition contrasts, and trained or stochastic systems should use at least three seeds. The random seed is not the unit of resampling; the source question remains the independent sampling unit.

\subsection{Partial-evidence necessity tests}

For two-support-paragraph instances, evaluate four inputs: both gold paragraphs, bridge only, answer-hop only, and neither. Define the complete-chain gain
\begin{equation}
G_{\mathrm{chain}}=M(q,\{b,t\})-
\max\{M(q,\{b\}),M(q,\{t\}),M(q,\varnothing)\}.
\label{eq:chain-gain}
\end{equation}
A positive value is evidence that the model benefits from both hops, although it is not a proof of faithful internal reasoning. Reporting $G_{\mathrm{chain}}$ by $(\ell_q,\ell_b,\ell_t)$ directly addresses whether language crossing changes evidence use rather than only final accuracy.

\section{Recovered prompts and evaluated model identifiers}\label{app:prompts}

\subsection{Scope and provenance}

Prompts are reported because small wording and message-assembly differences can affect multilingual generation. They do not by themselves make an experiment bitwise reproducible: model revision, demonstration choice, decoding, and parser behavior also matter. Table~\ref{tab:prompt-provenance} distinguishes text recovered literally from dependencies that were not frozen. Credential-bearing legacy scripts are excluded from every release; the boxes below contain sanitized prompt text only.

\begin{table}[htbp]
\caption{Provenance of the prompt contracts used in construction and evaluation.}
\label{tab:prompt-provenance}
\small
\begin{tabularx}{\linewidth}{@{}lYY@{}}
\toprule
Protocol & Recovered & Not fully recoverable \\
\midrule
V1 translation & Role-specific system prompts, user assembly, decoding & Provider-resolved revision; assignment seed \\
V2 RC1 translation & Versioned system prompt, response contract, retry behavior & Provider-side checkpoint identity \\
V1 readers & Instruction, one-shot row, message order & Endpoint revision for every historical run \\
Adapted selector & System/user messages and output target & Tokenizer and fine-tuned checkpoint revision \\
Translation judge & Unit-specific prompts, sampling design, parser, decoding & Checkpoint behind the requested GLM-5.2 alias \\
\bottomrule
\end{tabularx}
\end{table}

\subsection{Translation prompts}

The historical V1 generator used four role-specific system messages. The placeholder is replaced by the target-language name.

\begin{promptbox}{V1 title translation --- system message}
You are translator. Translate the TitleName from English to <target-language> do not use any other word, the output must be only translation
\end{promptbox}

\begin{promptbox}{V1 paragraph translation --- system message}
You are translator. Translate the Paragraph from English to <target-language> do not use any other word, the output must be only translation and your output style like input:\par
0. translate sentence0\par
1. translate sentence1\par
2. translate sentence2\par
...
\end{promptbox}

\begin{promptbox}{V1 question translation --- system message}
You are translator. Translate the Question from English to <target-language> do not use any other word, the output must be only translation
\end{promptbox}

\begin{promptbox}{V1 answer translation --- system message}
You are translator. Translate the Answer from English to <target-language> do not use any other word, the output must be only translation
\end{promptbox}

Title, question, and answer requests supplied only the English source string. Paragraph requests numbered the source sentences and expected the same numbered layout. V1 used the mutable \texttt{gpt-4o-mini} alias; its provider-resolved revision and the random assignment state were not retained, so the recovered contract is exact at the message level but not at the endpoint level.

V2 RC1 used the following stricter translation instruction with Gemma~4~31B.

\begin{promptbox}{V2 translation --- system message}
You are the deterministic translation component of a multilingual QA dataset. Preserve named entities, numbers, dates, yes/no polarity, and sentence boundaries. Do not answer the question and do not add explanations. The user request contains a response\_schema; return exactly one valid JSON object that satisfies it, with no additional keys and no Markdown.
\end{promptbox}

For a title, question, or answer, the response contract requires one non-empty \texttt{translation} value. For a paragraph, it requires a \texttt{translations} array with exactly the same number of non-empty elements as the ordered source-sentence array. The parser rejects prose, code fences, extra or duplicate keys, multiple JSON values, empty translations, and changed sentence cardinality. Exhausted requests are recorded as failures rather than silently copied from English.

\subsection{Source-referenced GLM judge prompts}\label{app:judge-prompts}

Paragraphs and questions share the following system message.

\begin{promptbox}{Translation judge for paragraphs and questions --- system message}
You are a meticulous bilingual translation judge. Your task is to assign a single integer score from 0 to 100 (0 = unusable, 100 = excellent) for a candidate translation, using only the source text and the target candidate.\par
Scoring rubric (decide internally; do not print sub-scores):\par
Adequacy/Faithfulness (60\%): Candidate conveys all meaning from the source; no omissions/additions; no contradictions. Use your bilingual understanding to compare source meaning with the target text.\par
Terminology, Entities, Numbers (15\%): Names, numbers, dates, units, and placeholders preserved and correct.\par
Fluency/Grammar (20\%): Natural, grammatical writing in the target language.\par
Style/Register (5\%): Tone/register appropriate to the source intent and genre.\par
Critical error floors:\par
Wrong language or untranslated copy of source -> <=10\par
Direct contradiction/mistranslation of key meaning -> <=40\par
Hallucinated content not grounded in source -> <=40\par
Loss/corruption of critical numbers, names, or placeholders -> <=60\par
Guidance: Favor semantic fidelity to the source over cleverness. Do not invent facts. Minor orthographic or punctuation quirks are minor unless they change meaning.\par
Round to the nearest integer and clamp to [0,100]. If inputs are empty or unusable, return 0.\par
Output format: respond with a short explanation (1--3 sentences) of the strengths and weaknesses of the candidate translation, then on a new final line output exactly ``SCORE: <integer>'' where <integer> is the score from 0 to 100.
\end{promptbox}

Answers use a separate instruction because a short entity, number, date, or yes/no response has different preservation criteria.

\begin{promptbox}{Translation judge for short answers --- system message}
You are a meticulous bilingual translation judge for short answers. You are given an English question (``context\_question'') and its short English answer (``text'') alongside a translated answer (``model\_output''). Your task is to judge how well ``model\_output'' translates the English answer into the target language.\par
Keep in mind: the answer is a concise factual response (a name, date, number, place, or yes/no). The translation must preserve the same factual content as the English answer, in the target language, while remaining natural.\par
Scoring rubric (decide internally; do not print sub-scores):\par
Faithfulness (70\%): The translated answer conveys the same fact(s) as the English answer. For yes/no answers the polarity must match.\par
Terminology, Entities, Numbers (20\%): Names, numbers, dates, and entities preserved and correct.\par
Fluency (10\%): Natural, grammatical form in the target language for a short answer.\par
Critical error floors:\par
Wrong language or untranslated English copy -> <=10\par
Opposite yes/no polarity or completely different fact -> <=20\par
Hallucinated content not grounded in the English answer -> <=40\par
Loss/corruption of critical numbers, names, or dates -> <=60\par
Guidance: Names that are conventionally kept in Latin script in the target language should not be penalized. Minor transliteration differences are acceptable.\par
Round to the nearest integer and clamp to [0,100]. If inputs are empty or unusable, return 0.\par
Output format: respond with a short explanation (1--3 sentences) of the strengths and weaknesses of the candidate answer translation, then on a new final line output exactly ``SCORE: <integer>'' where <integer> is the score from 0 to 100.
\end{promptbox}

The historical label ``Critical error floors'' is imprecise: because every rule specifies a maximum score, the rules act as ceilings. We retain the literal wording for reproducibility and use the correct term in the analysis. Each judge request supplies the English source, assigned target language, and candidate translation; answer requests additionally supply the English question. The requested model alias was \texttt{glm-5.2}, temperature was zero, and the parser read the integer from the final \texttt{SCORE:} line. The audit yields one source-referenced score, not a second reference-free score.

\subsection{Reader and selector prompts}

Table~\ref{tab:model-identifiers} organizes the open-model repository identifiers recorded for the historical reader screen. The exact checkpoint behind the endpoint historically labeled Llama~3.1~70B could not be recovered and is therefore treated as an artifact label in the article.

\begin{table}[htbp]
\caption{Recorded repository identifiers for the historical reader screen.}
\label{tab:model-identifiers}
\small
\begin{tabularx}{\linewidth}{@{}lY@{}}
\toprule
Family & Repository identifiers \\
\midrule
Llama 3.1 Instruct & \path{meta-llama/Meta-Llama-3.1-8B-Instruct}; \path{meta-llama/Meta-Llama-3.1-70B-Instruct}; \path{meta-llama/Meta-Llama-3.1-405B-Instruct} \\
Qwen2 Instruct & \path{Qwen/Qwen2-7B-Instruct}; \path{Qwen/Qwen2-72B-Instruct} \\
Aya 23 & \path{CohereForAI/aya-23-8B}; \path{CohereForAI/aya-23-35B} \\
Gemma 2 Instruct & \path{google/gemma-2-9b-it}; \path{google/gemma-2-27b-it} \\
\bottomrule
\end{tabularx}
\end{table}

\begin{promptbox}{Cross-lingual reader --- recovered system prefix}
You are a cross-lingual QA test machine. Your task is to answer the [Question] accurately based on the provided [Context], which may include multiple languages. Ensure that the answer is provided in the same language as the question. Produce only the correct answer without any additional text. (Hint: Some questions may require you to trace connections between components or entities across different languages in the context. To solve these, carefully piece together information from various sections, ensuring consistency with the question's language for a comprehensive answer.
\end{promptbox}

The complete one-shot reader variants append one preserved demonstration inside the system message, followed by separate user messages for the evaluated question and context. The Llama/GPT artifacts use demonstration row~2, while Qwen2~72B uses row~0; minor punctuation and whitespace differences are retained as distinct historical variants.

\begin{promptbox}{Cross-lingual evidence selector --- system message}
As a cross-lingual retriever, your task is to address a multi-hop question that involves navigating through 'information\_sources' available in multiple languages. Your goal is to use the 'question' as a directive to identify and extract information across different languages that collectively contribute to the answer. Multi-hop questions in this context require you to connect and synthesize data points across various languages, ensuring that the extracted information is relevant, accurate, and directly supports the answer. Focus on identifying the exact paragraphs and sentence indices (supporting facts) that provide necessary evidence to answer the question.
\end{promptbox}

The selector receives the serialized question and candidate set as one user message and predicts supporting paragraph/sentence identifiers. This text describes the historical selector; the released evaluator uses stable identifiers rather than depending on translated-title string matching.

\subsection{Rationale-teacher prompt}

\begin{promptbox}{Rationale teacher --- system message}
You are an advanced reasoning assistant capable of generating clear, logical, and step-by-step explanations for questions based on provided context. Given a Question, Context, and Answer, your task is to create a detailed 'Chain of Thought' (CoT) that explains how the answer is derived from the context. Ensure that your explanation is thorough, well-structured, and directly connects the information from the context to the final answer.
\end{promptbox}

The next three messages contain the question, context, and gold answer. The recovered supervised-reader serialization uses \texttt{Question:}, \texttt{Context:}, and \texttt{answer: ***<answer>***}; rationale targets append the teacher explanation. The provider-resolved teacher revision and complete tokenizer/trainer state were not retained, so we report this as a recovered historical protocol rather than a bitwise reconstruction.

\ifdefined\XHotpotQAArchive
\else

\bibliographystyle{unsrtnat}
\bibliography{references}
\end{document}